\documentclass{article}

\usepackage[final]{neurips_2025}

\usepackage[utf8]{inputenc} 
\usepackage[T1]{fontenc}    
\usepackage{hyperref}       
\usepackage{url}            
\usepackage{booktabs}       
\usepackage{amsfonts}       
\usepackage{nicefrac}       
\usepackage{microtype}      
\usepackage{xcolor}         

\usepackage{amsmath}
\usepackage[textsize=tiny]{todonotes}
\usepackage{cleveref}
\usepackage{enumitem}
\usepackage{multirow}
\usepackage{graphicx}
\usepackage{wrapfig}
\usepackage{array}
\usepackage{amssymb}
\usepackage{pifont}
\usepackage{bbm}
\usepackage[table,xcdraw]{xcolor}
\usepackage{titletoc} 
\usepackage{algorithm}
\usepackage{algpseudocode}

\newcommand{\diff}{\text{d}}

\newcommand{\ldriftcoef}{\mathbf{F}_t\mathbf{x}_t}

\newcommand{\ldiffcoef}{\mathbf{G}_t}
\newcommand{\wiener}{\mathbf{w}_t}
\newcommand{\rwiener}{\overline{\mathbf{w}}_t}
\newcommand{\score}{\nabla_{\mathbf{x}_t}{\log{p(\mathbf{x}_t)}}}
\newcommand{\scorec}{\nabla_{\mathbf{x}_t}{\log{p(\mathbf{x}_t\mid\mathbf{x}_0)}}}
\newcommand{\scorenet}{\mathbf{s}_{\boldsymbol{\theta}}(\mathbf{x}_t, t)}
\newcommand{\expval}{\mathbb{E}}
\newcommand{\ie}{\emph{i.e.}}
\newcommand{\eg}{\emph{e.g.}}
\newcommand{\cmark}{\ding{51}}  
\newcommand{\xmark}{\ding{55}} 

\newtheorem{theorem}{Theorem}

\newtheorem{proposition}{Proposition}
\newtheorem{lemma}{Lemma}

\title{System-Embedded Diffusion Bridge Models}

\author{%
  Bartlomiej Sobieski\thanks{Corresponding author at \texttt{b.sobieski@uw.edu.pl}} \hspace{0.1em}\thanks{Equal contribution} \hspace{0.1em}$^1$, Matthew Tivnan\footnotemark[2] \hspace{0.1em}$^{2,3}$, Yuang Wang$^{2,3}$, Siyeop Yoon$^{2,3}$, \\ \textbf{Pengfei Jin$^{2,3}$, Dufan Wu$^{2,3}$, Quanzheng Li$^{2,3}$, Przemyslaw Biecek$^{1,4}$} \\
  $^1$University of Warsaw, $^2$Harvard University, \\
  $^3$Massachusetts General Hospital, $^4$Warsaw University of Technology\\
}

\begin{document}

\maketitle

\begin{abstract}
  Solving inverse problems—recovering signals from incomplete or noisy measurements—is fundamental in science and engineering. Score-based generative models (SGMs) have recently emerged as a powerful framework for this task. Two main paradigms have formed: unsupervised approaches that adapt pretrained generative models to inverse problems, and supervised bridge methods that train stochastic processes conditioned on paired clean and corrupted data. While the former typically assume knowledge of the measurement model, the latter have largely overlooked this structural information. We introduce System-embedded Diffusion Bridge Models (SDBs), a new class of supervised bridge methods that explicitly embed the known linear measurement system into the coefficients of a matrix-valued SDE. This principled integration yields consistent improvements across diverse linear inverse problems and demonstrates robust generalization under system misspecification between training and deployment, offering a promising solution to real-world applications.
\end{abstract}

\section{Introduction}
\label{sec:introduction}

Restoring data from corrupted or incomplete measurements is a fundamental task in science and engineering \citep{PhysRevLett.116.061102,Akiyama_2019}, commonly referred to as an \emph{inverse problem}. Its \emph{linear} formulation plays a central role in practical domains such as signal processing and medical imaging \citep{M_Bertero_1985,1580791}. The emergence of deep learning has significantly advanced this field, enabling major scientific breakthroughs \citep{9084378}. Since the work of \cite{songscore}, the scientific community has increasingly adopted \emph{score-based generative models} (SGMs), also known as \emph{diffusion models}, to tackle inverse problems. Two key directions have emerged: one adapts models pretrained for image synthesis to conditional generation; the other, often called \emph{bridge methods}, trains problem-specific models grounded in stochastic differential equations (SDEs), assuming access to paired samples of clean data and corresponding measurements.

While pretrained models often assume access to a known linear measurement system, bridge methods have primarily focused on developing general-purpose approaches without leveraging such structural information. However, in many real-world settings—such as CT or MRI reconstruction—the linear measurement process is known \emph{a priori}, and datasets frequently contain paired examples of clean and degraded data. 

To address this gap, we propose \emph{System-embedded Diffusion Bridge Models} (SDBs), a novel method that incorporates the system response and noise covariance directly into the coefficients of a matrix-valued SDE. By embedding this measurement-system knowledge, SDB achieves considerable performance gains across diverse linear systems of varying complexity, demonstrated through three distinct instantiations. Furthermore, we conduct an extensive study of SDB's generalization under system misspecification between training and deployment, showing that SDB consistently outperforms baselines and remains robust in the face of parameter shifts, making it well-suited for real-world deployment.

In addition, our work advocates for more expressive formulations of diffusion processes in generative modeling, opening avenues for future research into system-aware and structure-driven stochastic modeling frameworks.\footnote{We include the source code at \href{https://github.com/sobieskibj/sdb}{https://github.com/sobieskibj/sdb}.}

\begin{figure}[t]
    \centering
\includegraphics[width=1.0\linewidth]{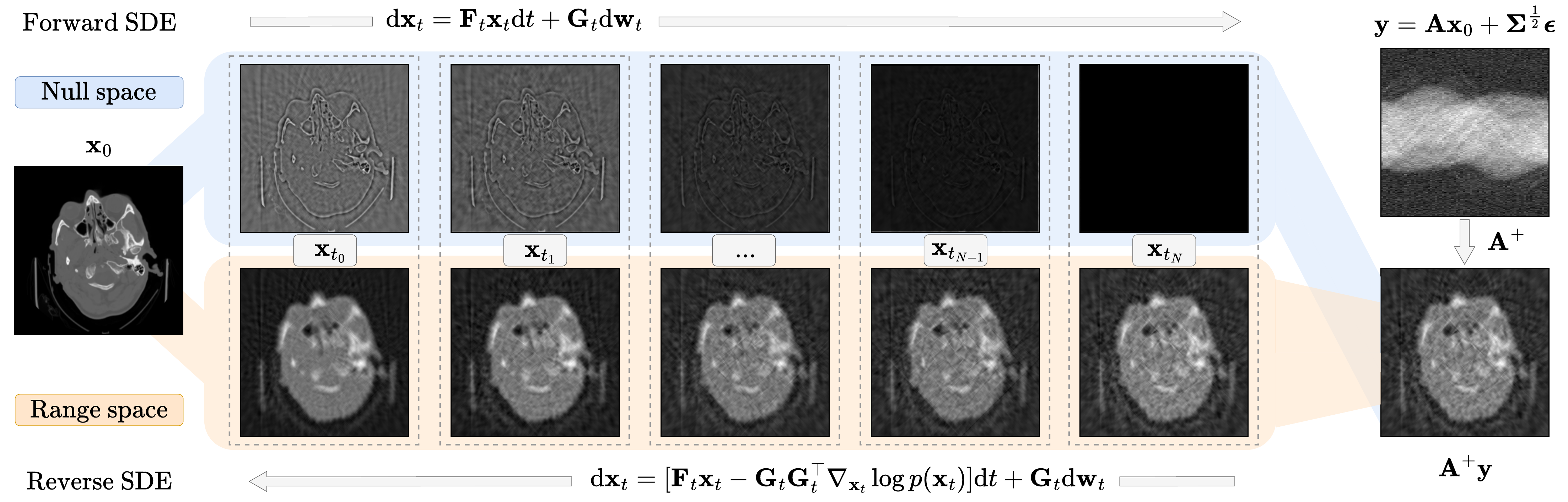}
    \caption{SDB learns a diffusion bridge between pseudoinverse reconstructions and clean samples by embedding the measurement system into the coefficients of the linear SDE, allowing the score model to distinguish between the range and null spaces of the task.}
    \label{fig:teaser}
\end{figure}

\section{Related works}
\label{sec:related_works}
Our work offers a new perspective on constructing task-specific diffusion bridges (\emph{supervised}), while drawing inspiration from diffusion-based plug-and-play methods (\emph{unsupervised}). Both directions address the topic of image restoration and inverse problems from different viewpoints. Below, we briefly cover these directions, with a much more detailed overview included in the Appendix.

\paragraph{Diffusion models for inverse problems.} As shown in the work of \citet{songscore}, pretrained unconditional diffusion models can be easily adapted to conditional synthesis with a simple application of Bayes' Theorem. The resulting likelihood term has been extensively utilized for guided generation \citep{dhariwal2021beat}, especially for image restoration problems \citep{daras2024survey}. These methods range from generic ones \citep{songscore,chung2022improving} to those that incorporate additional assumptions and constraints to the generation process, such as DPS \citep{chung2023diffusion}, DDNM \citep{wangzero}, $\Pi \text{GDM}$ \citep{song2023pseudoinverseguided}, DDPG \citep{Garber_2024_CVPR}, and others \citep{kawar2022denoising,chung2022improving, chung2023diffusion,song2023pseudoinverseguided, mardanivariational,chung2024decomposed}.

\paragraph{Diffusion bridges.} One may also extend the SGM framework to arrive at mappings between arbitrary probability distributions given paired data \citep{sarkka2019applied}. Among many seminal works in this area \citep{heng2021simulating,somnath2023aligned,peluchetti2022nondenoising,delbracio2023inversion,de2021diffusion}, \citet{liu20232} (I2SB) propose a simulation-free algorithm that solves the Schrödinger Bridge (SB) problem, \citet{luo2023image} (IR-SDE) construct a conditional stochastic differential equation (SDE) that maps noisy degradations to clean samples, while \citet{yue2024image} (GOUB) generalize it with the Doob's h-transform \citep{doob1984classical}, which is also used by \citet{zhoudenoising} (DDBM) to generalize the entire SGM framework. 

\section{Background}
\label{sec:background}

\paragraph{Diffusion models.}
Generative modeling of visual data has advanced rapidly with the introduction of diffusion models \citep{sohl2015deep, ho2020denoising, dhariwal2021beat}, which generate images by sequential denoising. Initially formulated as a discrete Markov chain with Gaussian kernels, diffusion models were later unified with score matching \citep{song2019generative} under the framework of SDEs \citep{songscore}, establishing the continuous-time formulation known as SGM.

Let $p(\mathbf{x}_t)$ denote the probability distribution of $\mathbf{x}_t \in \mathbb{R}^d$ parameterized by time $t \in [0, 1]$, where $p(\mathbf{x}_0)$ represents the data distribution and $p(\mathbf{x}_1)$ the prior. In this work, we highlight the often-overlooked \emph{linear} formulation of SGM. The forward SDE, which maps $p(\mathbf{x}_0)$ to $p(\mathbf{x}_1)$, is defined as
\begin{equation}
\diff \mathbf{x}_t = \ldriftcoef \diff t + \ldiffcoef \diff \wiener,\label{eq:linear_forward_sde}
\end{equation}
where $\ldriftcoef$ is a linear drift term with time-dependent matrix $\mathbf{F}_t \in \mathbb{R}^{d\times d}$, and $\ldiffcoef$ is a matrix-valued diffusion coefficient. Following \citet{anderson1982reverse}, the reverse SDE, which maps $p(\mathbf{x}_1)$ to $p(\mathbf{x}_0)$, is given by
\begin{equation}
\diff \mathbf{x}_t = [\ldriftcoef - \ldiffcoef \ldiffcoef^\top \score] \diff t + \ldiffcoef \diff \rwiener, \label{eq:linear_reverse_sde}
\end{equation}
where $\score$ is the score function and $\rwiener$ is the Wiener process with reversed time. The forward-reverse relationship ensures that the marginal distributions $p(\mathbf{x}_t)$, defined by \cref{eq:linear_forward_sde} and \cref{eq:linear_reverse_sde}, remain consistent.

In standard SGM for image synthesis, $p(\mathbf{x}_1)$ is typically chosen as an isotropic Gaussian. To generate samples from $p(\mathbf{x}_0)$, one samples from $p(\mathbf{x}_1)$ and solves \cref{eq:linear_reverse_sde} using a time discretization scheme. This requires the unknown score function $\score$, which is generally approximated by a neural network $\scorenet$ trained with the score-matching loss \citep{songscore}:
\begin{equation}
\expval [ | \scorenet - \scorec |^2_2 ],
\label{eq:score_loss}
\end{equation}
where the expectation is taken over $\mathbf{x}_0 \sim p(\mathbf{x}_0)$, $t \sim \mathcal{U}([0, 1])$, and $\mathbf{x}_t \sim p(\mathbf{x}_t \mid \mathbf{x}_0)$.

The linear formulation in \cref{eq:linear_forward_sde} and \cref{eq:linear_reverse_sde} is a special case of the broader SGM framework introduced by \citet{songscore}, where drift and diffusion coefficients can be arbitrarily complex functions of $\mathbf{x}_t$ and $t$. This linear formulation, however, generalizes the well-established \emph{scalar} case. Specifically, the Variance-Preserving (VP) SDE, which maintains constant variance for $\mathbf{x}_t$ across time, can be recovered with $\mathbf{F_t} = -\frac{1}{2} \beta_t \mathbf{I}$ and $\mathbf{G}_t = \sqrt{\beta_t} \mathbf{I}$. The Variance-Exploding (VE) SDE, which induces unbounded growth in the variance of $p(\mathbf{x}_1)$ as $t \to 1$, is obtained with $\mathbf{F_t} = \mathbf{0}$ and $\mathbf{G}_t = \sqrt{\frac{\diff \sigma^2_t}{\diff t}} \mathbf{I}$. While these scalar matrices offer simplicity, they also raise the question of whether more sophisticated designs could yield additional benefits.

\paragraph{Inverse problems.} For a signal sample $\mathbf{x}\in\mathbb{R}^d$, we define the \emph{linear measurement system} as
\begin{equation}
    \mathbf{y} = \mathbf{A}\mathbf{x} + \boldsymbol{\Sigma}^{\frac{1}{2}} \boldsymbol{\epsilon},
    \label{eq:measurement_system}
\end{equation}
where $\mathbf{y}\in\mathbb{R}^m$ is the \emph{measurement}, $\boldsymbol{A} \in \mathbb{R}^{m\times d}$ is the \emph{system response matrix}, $\boldsymbol{\epsilon}\sim\boldsymbol{\mathcal{N}}(\mathbf{0}_{m}, \mathbf{I}_{m\times m})$ is the measurement noise with covariance matrix $\boldsymbol{\Sigma}\in\mathbb{R}^{m\times m}$ and $p(\mathbf{y} \mid \mathbf{x} ) = \boldsymbol{\mathcal{N}}(\mathbf{A}\mathbf{x}, \boldsymbol{\Sigma})$ is the \emph{forward model}. We refer to finding the unknown $\mathbf{x}$ based on the provided $\mathbf{y}$ as solving a linear inverse problem with additive Gaussian noise \citep{tarantola2005inverse}. This formulation is central to many practical applications, such as signal processing, medical imaging, and remote sensing \citep{mueller2012linear}, particularly in the case of noninvertible linear systems, where either $m \neq d$ or $\mathbf{A}$ is rank-deficient.

\paragraph{Matrix pseudoinverse.}
Consider the simplest noiseless ($\boldsymbol{\Sigma} = \mathbf{0}_{m\times m}$) setting of \cref{eq:measurement_system} with invertible $\mathbf{A}$. In this case, the original signal $\mathbf{x}$ can be fully recovered by simply using the inverse of $\mathbf{A}$, \ie, $\mathbf{x}=\mathbf{A}^{-1}\mathbf{y}$. However, this is only possible when $m = d$ and $\mathbf{A}$ is full-rank, which greatly limits the scope of potential inverse problems. In practice, the matrix 
$\mathbf{A}$ is typically not invertible, and noise arises naturally due to, \eg,  imperfections in the measurement process.

In the case of non-invertible $\mathbf{A}$, one may utilize its Moore-Penrose inverse, often called the \emph{pseudoinverse}, denoted as $\mathbf{A}^+$. It generalizes the notion of standard inverse of square matrices to those of arbitrary shape and rank. Consider the singular value decomposition (SVD) $\mathbf{A} = \mathbf{U}\mathbf{D}\mathbf{V}^*$ of $\mathbf{A}$ for matrices $\mathbf{U}, \mathbf{D}, \mathbf{V}$, where $\mathbf{U}$ and $\mathbf{V}$ are unitary, and $\mathbf{D}$ is diagonal with non-negative real entries. Then, its pseudoinverse is computed via $\mathbf{A}^+ = \mathbf{V}\mathbf{D}^+\mathbf{U}^*$, where $\mathbf{D}^+$ is obtained by replacing all of its non-zero singular values with their reciprocals.

\paragraph{Range-nullspace decomposition.} A particularly useful property of the matrix pseudoinverse is its role in the well-known range-nullspace decomposition \citep{strang2022introduction}. Specifically, any signal $\mathbf{x}$ can be decomposed as
\begin{equation}
    \mathbf{x} = 
    \underbrace{\vphantom{(\mathbf{I} - \mathbf{A}^+\mathbf{A})}\mathbf{A}^+\mathbf{A}\mathbf{x}}_{\textit{range space}} + 
    \underbrace{(\mathbf{I} - \mathbf{A}^+\mathbf{A})\mathbf{x}}_{\textit{null space}},
    \label{eq:rn_decomposition}
\end{equation}
where the first term lies in the range (image) of $\mathbf{A}$, and the second in its null space (kernel), \ie, it is annihilated by $\mathbf{A}$. This decomposition is particularly relevant in inverse problems: applying $\mathbf{A}$ to $\mathbf{x}$ eliminates the null space component, while preserving the range component. Consequently, solving an inverse problem can be viewed as denoising in the range space and synthesizing missing information in the null space.

\paragraph{Pseudoinverse reconstruction.}
A matrix pseudoinverse can also be used to compute the \emph{pseudoinverse reconstruction} (PR) $\hat{\mathbf{x}}$ of $\mathbf{x}$ via $\hat{\mathbf{x}} = \mathbf{A}^+\mathbf{y}$. Crucially, this reconstruction always lies in $\mathbb{R}^d$, regardless of the shape of $\mathbf{y}$. This property addresses a common limitation of prior bridge methods, \ie, their default operation in fixed-dimensional spaces. When $m \neq d$, such methods require additional effort to remain well-defined. For more challenging inverse problems, such as those in medical imaging, PR provides a principled mechanism for resolving this issue.

From a theoretical perspective, assuming access to $\mathbf{A}^+$ also simplifies the relationship between $\mathbf{x}$ and $\mathbf{y}$ \citep{strang2022introduction}. When $m < d$, the PR is the unique minimizer of $\| \mathbf{A}\mathbf{x}^* - \mathbf{y} \|_2$ over all $\mathbf{x}^*\in\mathbb{R}^d$. When $m > d$, it yields the minimum $\ell_2$-norm solution among those satisfying $\mathbf{A}\mathbf{x}^* = \mathbf{y}$, \ie, $\min_{\mathbf{x}^*} \| \mathbf{x}^* \|_2 \quad \text{subject to} \quad \mathbf{A}\mathbf{x}^* = \mathbf{y}.$ These optimality properties further motivate the use of PR when $\mathbf{A}$ is known and its pseudoinverse is tractable.

\paragraph{Problem-specific diffusion bridges.} Many unsupervised approaches, assuming access to a specific noisy linear system (\cref{eq:measurement_system}), offer tailored solutions that improve performance under additional assumptions \citep{song2023pseudoinverseguided,wangzero,chung2023diffusion,Garber_2024_CVPR}. In contrast, state-of-the-art (SOTA) diffusion bridges \citep{liu20232,luo2023image,yue2024image,zhoudenoising} make no assumptions about the underlying system during training, focusing on general mappings between arbitrary distributions. This can be suboptimal when the system is (even approximately) known. For instance, in image inpainting, these methods do not distinguish between the known range (unmasked) and the missing null part (masked), resulting in redundant noise in the range space, which is expected to be noiseless. In more complex scenarios, such redundancy could reduce efficiency, while ignoring system-specific information may hinder generalization. Thus, developing a supervised diffusion bridge tailored to a specific system remains an important yet unresolved challenge.

\section{Method}
\label{sec:method}
In this section, we seek to construct a bridge diffusion process that directly incorporates the information about the assumed measurement system. By that, we understand a process which i) maps the PRs to the respective signal samples, ii) synthesizes the missing information directly in the null space and iii) optionally denoises the range space part. We begin by noticing a specific connection between how intermediate steps of a general linear diffusion process are obtained and how noisy linear measurement systems map clean samples to observations. As our key contribution, we propose a specific design of matrix-valued SDEs, which fulfill the initially desired properties.

\paragraph{Connecting systems to SDEs.}
Following \cref{eq:measurement_system}, it is clear that $\mathbf{y}\mid \mathbf{x}\sim \boldsymbol{\mathcal{N}}(\mathbf{A}\mathbf{x}, \boldsymbol{\Sigma})$, \ie, for a given signal $\mathbf{x}$, a noisy linear system applies a linear transformation $\mathbf{A}$ and adds (possibly correlated) Gaussian noise to sample a measurement. On the other hand, given a clean sample $\mathbf{x}_0$, \cref{eq:linear_forward_sde} transforms it to $\mathbf{x}_t$ by applying a \emph{cascade} of such noisy linear systems, which is a noisy linear system itself. That is, $\mathbf{x}_t \mid \mathbf{x}_0 \sim \boldsymbol{\mathcal{N}}(\mathbf{H}_t\mathbf{x}_0, \boldsymbol{\Sigma}_t)$ for some time-dependent matrices $\mathbf{H}_t, \boldsymbol{\Sigma}_t$. Hence, one may equivalently arrive at $\mathbf{y} \mid \mathbf{x}$ (or some linear transformation of $\mathbf{y}$) through an application of a series of noisy linear systems to $\mathbf{x}$ instead of just a single one. In the limit, this would imply the existence of a general linear SDE that performs such mapping.

The above considerations suggest that the measurement system could be directly \emph{embedded} into $\mathbf{H}_t$ and $\boldsymbol{\Sigma}_t$ to arrive at such SDE. While the relationship between these and the drift and diffusion coefficients of the corresponding stochastic process is not immediately clear, a recent theoretical result derived by \citet{10937272} provides a convenient way of mapping from one to the other under mild assumptions. We restate it below for clarity. 

\begin{theorem}\label{th:sde_to_cascade}
\citep{10937272} Assume that $\mathbf{x}_t \mid \mathbf{x}_0$ evolves according to the linear forward process from \cref{eq:linear_forward_sde}. Then, $\mathbf{x}_t \mid \mathbf{x}_0 \sim \mathbf{\mathcal{N}}(\mathbf{H}_t\mathbf{x}_0, \boldsymbol{\Sigma}_t)$, where
\begin{equation}
    \mathbf{H}_t = \exp{\left(\boldsymbol{\Omega}_t(\mathbf{F}_t)\right)}\approx\exp{\left(\int_0^t{\mathbf{F}_s \diff s}\right)}, \boldsymbol{\Sigma}_t = \mathbf{H}_t \left( \int_0^t \mathbf{H}_s^{-1} \mathbf{G}_s\mathbf{G}^\top_s\mathbf{H}_s^{-1^{\top}} \diff s\right) \mathbf{H}_t^\top,\label{eq:cascade}
\end{equation}
where $\boldsymbol{\Omega}_t$ is the Magnus expansion and the approximation becomes equality if, for all $s, s'\in[0, t]$, $[\mathbf{F}_s, \mathbf{F}_{s'}]=0$, \ie, $\mathbf{F}_s$ and $\mathbf{F}_{s'}$ commute. Conversely, given $\mathbf{H}_t$ and $\boldsymbol{\Sigma}_t$, the drift and diffusion coefficients can be obtained via:
\begin{equation}
    \mathbf{F}_t = \frac{\diff \mathbf{H}_t}{\diff t}\mathbf{H}_t^{-1}, \mathbf{G}_t \mathbf{G}_t^\top = \frac{\diff \boldsymbol{\Sigma}_t}{\diff t} - \mathbf{F}_t \boldsymbol{\Sigma}_t - \boldsymbol{\Sigma}_t \mathbf{F}_t^\top.\label{eq:inverse_cascade}
\end{equation}
\end{theorem}
Therefore, \cref{th:sde_to_cascade} allows one to obtain the linear diffusion process governing the evolution of $\mathbf{x_t}\mid\mathbf{x}_0$ defined through $\mathbf{H}_t$ and $\mathbf{\Sigma}_t$, which in turn makes training and sampling from continuous-time models possible.

\paragraph{Embedding the measurement system.}
Equipped with the necessary tools, we propose to embed the measurement system into the linear diffusion process by using
\begin{subequations} \label{eq:sdb_cascade}
\begin{align}
\mathbf{H}_t&=\mathbf{A}^+\mathbf{A} + \alpha_t(\mathbf{I} - \mathbf{A}^+\mathbf{A}),\label{eq:sdb_cascade_linear} \\
\boldsymbol{\Sigma}_t&=\gamma_t\mathbf{A}^+\boldsymbol{\Sigma}{\mathbf{A}^{+}}^\top + \beta_t(\mathbf{I} - \mathbf{A}^+\mathbf{A}).\label{eq:sdb_cascade_covar}
\end{align}
\end{subequations}
for $\alpha_t, \beta_t, \gamma_t$ being the null space drift, null space diffusion and range space diffusion coefficients respectively. We refer to the process resulting from \cref{eq:sdb_cascade_linear,eq:sdb_cascade_covar} as the \emph{System-embedded Diffusion Bridge Model} (SDB).

To better understand the rationale behind this specific design, it is worth considering the range and null space components of the resulting $\mathbf{x}_t$ separately. Specifically,
\begin{subequations} \label{eq:sdb_range_null}
\begin{align}
\mathbf{A}^+\mathbf{A}\mathbf{x}_t &= \mathbf{A}^+(\mathbf{A}\mathbf{x}_0 + \gamma_t\boldsymbol{\Sigma}^{\frac{1}{2}}\boldsymbol{\epsilon})=\mathbf{A}^+\mathbf{A}\mathbf{x}_0 + \sqrt{\gamma_t}\mathbf{A}^+\boldsymbol{\Sigma}^{\frac{1}{2}}\boldsymbol{\epsilon},\label{eq:sdb_range} \\
(\mathbf{I} - \mathbf{A}^+\mathbf{A})\mathbf{x}_t&=\alpha_t(\mathbf{I} - \mathbf{A}^+\mathbf{A})\mathbf{x}_0 + \sqrt{\beta_t}(\mathbf{I} - \mathbf{A}^+\mathbf{A})\boldsymbol{\epsilon}',\label{eq:sdb_null}
\end{align}
\end{subequations}
where $\boldsymbol{\epsilon} \sim \boldsymbol{\mathcal{N}}(\mathbf{0}_m, \boldsymbol{\Sigma}_{m\times m}), \boldsymbol{\epsilon}'\sim\boldsymbol{\mathcal{N}}(\mathbf{0}_d, \boldsymbol{\Sigma}_{d\times d})$. The range space part contains the original signal $\mathbf{A}^+\mathbf{A}\mathbf{x}_0$ and the stochastic component $\gamma_t\mathbf{A}^+\boldsymbol{\Sigma}^{\frac{1}{2}}\boldsymbol{\epsilon}$, which directly models the range space noise. If $\boldsymbol{\Sigma}=\mathbf{0}$, the true signal is fully recovered at every timestep $t$. The null space part is a mixture of the null component of the true signal $\alpha_t(\mathbf{I} - \mathbf{A}^+\mathbf{A})\mathbf{x}_0$ and null-space-projected Gaussian noise $\beta_t(\mathbf{I} -\mathbf{A}^+\mathbf{A})\boldsymbol{\epsilon}'$. Notably, SDB links two stochastic processes that evolve simultaneously within the range and null spaces, each modeled with independent noise variables, $\boldsymbol{\epsilon}$ and $\boldsymbol{\epsilon}'$. Moreover, it is evident that proper choices for $\alpha_t,\beta_t$ and $\gamma_t$ lead to a mapping between the PRs at $t=1$ to their respective clean samples at $t=0$.

\paragraph{SDE perspective.} Applying \cref{th:sde_to_cascade} to the coefficients from \cref{eq:sdb_cascade} leads to the following drift and diffusion coefficients for SDB:
\begin{subequations} \label{eq:sdb_drift_diffusion}
\begin{align}
\mathbf{F}_t&=\frac{\diff}{\diff t}\log{\alpha_t}(\mathbf{I} - \mathbf{A}^+\mathbf{A}),\label{eq:sdb_drift} \\
\mathbf{G}_t\mathbf{G}_t^\top&=\frac{\diff\gamma_t}{dt}\mathbf{A}^+\boldsymbol{\Sigma}\mathbf{A}^{+^{\top}} + \left( \frac{\diff\beta_t}{\diff t}-2\beta_t\frac{\diff}{\diff t}\log{\alpha_t}\right)(\mathbf{I} - \mathbf{A}^+\mathbf{A}).\label{eq:sdb_diffusion}
\end{align}
\end{subequations}
With this formulation at hand, we now propose a specific setting of the scalar functions $\alpha_t$ and $\beta_t$, which extend the result of \cite{liu20232} and make direct connection of SDB to the optimal transport (OT) plan \citep{mikami2004monge}.
\begin{theorem}
For the linear SDE defined in \cref{eq:linear_forward_sde} with $\mathbf{F}_t$ and $\mathbf{G}_t$ given by \cref{eq:sdb_drift,eq:sdb_diffusion} respectively, let $\alpha_t=\frac{\bar{\sigma}_t^2}{\bar{\sigma}_t^2 + \sigma_t^2},\beta_t=\frac{\sigma_t^2\bar{\sigma}_t^2}{\bar{\sigma}_t^2 + \sigma_t^2}$ for $\sigma_t^2=\int_0^t g^2(\tau)d\tau, \bar{\sigma}_t^2=\int_t^1 g^2(\tau)d\tau$ for some non-negative function $g(t)$. Assuming that $g(t) \rightarrow 0$ uniformly for all $t$, the null space part of this SDE reduces to an OT-ODE:
\begin{equation}
    \diff (\mathbf{I} - \mathbf{A}^+\mathbf{A})\mathbf{x}_t = \mathbf{v}_t(\mathbf{x}_t \mid \mathbf{x}_0) \diff t,
\end{equation}
where $\mathbf{v}_t(\mathbf{x}_t \mid \mathbf{x}_0)=\left( \lim_{g(t)\rightarrow 0}\frac{g^2(t)}{\sigma_t^2}\right)(\mathbf{I} - \mathbf{A}^+\mathbf{A})(\mathbf{x}_t - \mathbf{x}_0).$
\label{th:ot_ode}
\end{theorem}
We include the proof in the Appendix. To align the dynamics between the range and null space, we also use $\gamma_t=\frac{\sigma_t^2}{\bar{\sigma}_t^2 + \sigma_t^2}$. In practice, we do not set $g(t)=0$, but rather keep it at sufficiently low values. Because of the relationship to the SB problem, we term this variant as SDB (SB).

\paragraph{Novel processes.} To showcase the versatility of our framework, we introduce two additional variants of SDB that reinterpret the VP and VE diffusion processes of \citet{songscore}:
\begin{description}[labelindent=1em,leftmargin=0em]
    \item[SDB (VP):] $\alpha_t=1-t, \beta_t=\sqrt{1 - \alpha_t}, \gamma_t=\beta_t$,
    \item[SDB (VE):] $\alpha_t=1, \beta_t=\sigma_{max}\sqrt{t},\gamma_t=\sqrt{t}$, where $\sigma_{max}\gg1$.
\end{description}
In both cases, the original VP or VE process is applied in the null space, while the range space coefficient $\gamma_t$ is related to $\beta_t$ to simplify the dynamics. Unlike methods such as DDBM, which symmetrize the variance schedule around $t = 0.5$, these variants explicitly control how the original signal is erased in the null space. There, SDB (VP) performs a convex interpolation between the clean input and Gaussian noise, while SDB (VE) converges to an isotropic Gaussian $\boldsymbol{\mathcal{N}}(\mathbf{0}, \sigma_{\text{max}}^2 \mathbf{I})$ as $t \to 1$. As such, both more closely resemble approaches like IR-SDE that retain non-singular marginals at the endpoint. For simplicity, we parameterize SDB (VP) and SDB (VE) processes with linear scheduling, leaving more sophisticated designs as future work.

\paragraph{Principled posterior sampling.}
In the asymptotic limit of infinite data and model capacity, it is natural to ask whether SDB constitutes an exact probabilistic model of the underlying inverse problem—that is, whether it produces \emph{principled posterior samples} from $p(\mathbf{x}|\mathbf{y})$. The following result provides a positive answer under mild conditions on the forward process coefficients, with the proof deferred to the Appendix.

\begin{theorem}\label{th:principled_posterior_sampler}
Let the forward measurement model be $p(\mathbf{y}|\mathbf{x}) = \mathcal{N}(\mathbf{A}\mathbf{x}, \boldsymbol{\Sigma})$. 
Then, under the SDB dynamics defined by \cref{eq:linear_forward_sde}, \cref{eq:inverse_cascade}, and \cref{eq:sdb_cascade}, 
with time-dependent scalar coefficients $\alpha_t$, $\beta_t$, and $\gamma_t$ satisfying
\[
\lim_{t \rightarrow 1} \gamma_t = 1, 
\quad 
\lim_{t \rightarrow 1} \frac{\alpha_t^2}{\beta_t} = 0,
\]
the corresponding reverse-time SDE generates asymptotically exact samples from the Bayesian posterior distribution $p(\mathbf{x}|\mathbf{y})$.
\end{theorem}

\begin{table}[h]
\centering
\fontsize{27}{29}\selectfont
\renewcommand{\arraystretch}{2.5}
\caption{Comparison of prior SOTA diffusion bridge methods with SDB. Subsequent rows denote the drift ($\mathbf{F}_t$) and diffusion ($\mathbf{G}_t$) coefficients of the forward process, while $\boldsymbol{\mu}_t,\boldsymbol{\Sigma}_t$ are its respective mean and covariance at timestep $t$. Each column follows the original notation of each paper.}
\resizebox{\linewidth}{!}{%
\begin{tabular}{lccccc}
\toprule
\textbf{Method} & \textbf{I2SB} & \textbf{IR-SDE} & \textbf{GOUB} & \textbf{DDBM (VP)} & \textbf{SDB (SB)} \\
\midrule
$\mathbf{F}_t$ & \shortstack{$\frac{\beta_t}{\bar{\sigma}_2^t + \sigma_2^t}(\mathbf{x_1} - \mathbf{x}_0)-$\\$\beta_t\frac{\bar{\sigma}_2^t - \sigma_2^t}{\sigma_t^2\bar{\sigma}_2^t}(\mathbf{x}_t - \boldsymbol{\mu}_t)$} & $\theta_t(\mathbf{x}_1 - \mathbf{x}_t)$ & $(\theta_t + g^2(t) \frac{e^{-2 \bar{\theta}_{t:1}}}{\bar{\sigma}_{t:1}^{2}})(\mathbf{x}_1 - \mathbf{x}_t)$ & \shortstack{$(\frac{\diff}{\diff t}\log{\alpha_t})\mathbf{x}_t +$\\$ g^2(t)\frac{ \left( \frac{\alpha_t}{\alpha_1} \mathbf{x}_1 - \mathbf{x}_t \right) }{ \sigma_t^2 \left( \frac{\mathrm{SNR}_t}{\mathrm{SNR}_1} - 1 \right) }
$} & $\frac{\diff}{\diff t}\log{\alpha_t}(\mathbf{I} - \mathbf{A}^+\mathbf{A})$\\
\hline
$\mathbf{G}_t\mathbf{G}_t^\top$ & $\beta_t\mathbf{I}$ & $\sigma^2(t)\mathbf{I}$ & $g^2(t)\mathbf{I}$ & $g^2(t)\mathbf{I}$ & 
\shortstack{$\frac{\diff\gamma_t}{dt}\mathbf{A}^+\boldsymbol{\Sigma}\mathbf{A}^{+^\top}+$\\$ \left( \frac{\diff\beta_t}{\diff t}-2\beta_t\frac{\diff}{\diff t}\log{\alpha_t}\right)(\mathbf{I} - \mathbf{A}^+\mathbf{A})$} \\
\hline
$\boldsymbol{\mu}_t$ & 
\shortstack{$\frac{\bar{\sigma}^2_t}{\bar{\sigma}^2_t + \sigma_t^2}\mathbf{x}_0 +$\\$ \frac{\sigma^2_t}{\sigma^2_t + \sigma_t^2}\mathbf{x}_1$} & 
\shortstack{$ \mathbf{x}_1+$\\$(\mathbf{x}_0 - \mathbf{x}_1)e^{-\bar{\theta}_{0:t}}$} & 
\shortstack{$e^{-\bar{\theta}_{0:t}}\frac{\bar{\sigma}_{t:1}^2}{\bar{\sigma}_{0:1}^2}\mathbf{x}_0 +$\\$[(1 - e^{-\bar{\theta}_{0:t}})\frac{\bar{\sigma}_{t:1}^2}{\bar{\sigma}_{0:1}^2} +$\\$ e^{-2\bar{\theta}_{t:1}}\frac{\bar{\sigma}_{0:t}^2}{\bar{\sigma}_{0:1}^2}]\mathbf{x}_1$} & \shortstack{$\frac{\text{SNR}_1}{\text{SNR}_t}\frac{\alpha_t}{\alpha_1}\mathbf{x}_1 +$\\$\alpha_t(1 - \frac{\text{SNR}_1}{\text{SNR}_t})\mathbf{x}_0$} & \shortstack{$\mathbf{A}^+\mathbf{A}\mathbf{x}_0 + \alpha_t(\mathbf{I} - \mathbf{A}^+\mathbf{A})\mathbf{x}_0$}  \\
\hline
$\boldsymbol{\Sigma}_t$ & $\frac{\bar{\sigma}^2_t \sigma^2_t}{\bar{\sigma}^2_t + \sigma_t^2}\mathbf{I}$ & $\lambda^2(1 - e^{-2\bar{\theta}_t})\mathbf{I}$ & $\frac{\bar{\sigma}_{0:t}^2\sigma_{t:1}^2}{\bar{\sigma}_{0:1}^2}\mathbf{I}$ & $\sigma_t^2(1 - \frac{\text{SNR}_1}{\text{SNR}_t})\mathbf{I}$ & \shortstack{$\gamma_t\mathbf{A}^+\boldsymbol{\Sigma}{\mathbf{A}^{+}}^\top + $\\$\beta_t(\mathbf{I} - \mathbf{A}^+\mathbf{A})$\vspace{-0.5em}}\\
\hline
Markovian & \xmark & \xmark & \xmark & \xmark & \cmark \\
\hline
Hyperparameters & 
\shortstack{$\beta_t,$ \\ $\sigma_t^2 = \int_0^t{g^2(\tau)d\tau},$ \\ $\bar{\sigma}_t^2 = \int_t^1{g^2(\tau)d\tau}$} & 
\shortstack{$\theta_t=\frac{\sigma^2_t}{\lambda^2},$ \\ $\bar{\theta}_t=\int_0^t{\theta_\tau}d\tau$} & 
\shortstack{$\theta_t=\frac{g^2(t)}{2\lambda^2},$\\$\bar{\theta}_{s:t}=\int_s^t\theta_\tau d\tau,$\\$\bar{\sigma}_{s:t}^2=\frac{g^2(t)}{2\theta_t}(1 - e^{-2\bar{\theta}_{s:t}})$} & \shortstack{$\text{SNR}_t=\frac{\alpha_t^2}{\sigma_t^2},$\\$\alpha_t=\exp{\left(-\frac{1}{2}\int_0^t{g^2(\tau)d\tau}\right)},$\\$\sigma_t^2=1-\alpha_t^2$} & \shortstack{$\alpha_t, \beta_t, \gamma_t$\\follow from \cref{th:ot_ode}}\\
\bottomrule
\end{tabular}
}
\label{tab:comparison}
\end{table}

\section{Experiments}
\label{sec:experiments}

\begin{algorithm}
\caption{SDB Training}
\label{alg:sdb_training}
\begin{algorithmic}[1]
\Require $p(\mathbf{x}_0), p(t), \mathbf{A}, \mathbf{A}^+, \boldsymbol{\Sigma}^{1/2}, \alpha_t, \beta_t, \gamma_t, \mathcal{\boldsymbol{D}}_{\boldsymbol{\theta}}$
\For{each iteration}
    \State $\mathbf{x} \sim p(\mathbf{x}_0)$ \Comment{sample clean data $\mathbf{x}$}
    \State $t \sim p(t)$ \Comment{sample diffusion timestep $t$}
    \State $\boldsymbol{\epsilon} \sim \mathcal{N}(\mathbf{0}, \mathbf{I}) \in \mathbb{R}^{m}$ \Comment{sample range-space Gaussian noise $\boldsymbol{\epsilon}$}
    \State $\boldsymbol{\epsilon}' \sim \mathcal{N}(\mathbf{0}, \mathbf{I}) \in \mathbb{R}^{d}$ \Comment{sample null-space Gaussian noise $\boldsymbol{\epsilon}'$}
    \State $\mathbf{x}_t \gets [\mathbf{A}^+\mathbf{A} + \alpha_t(\mathbf{I} - \mathbf{A}^+\mathbf{A})] \mathbf{x} + \gamma_t^{\frac{1}{2}}\mathbf{A}^+\boldsymbol{\Sigma}^{\frac{1}{2}}\boldsymbol{\epsilon} + \beta_t^{\frac{1}{2}}(\mathbf{I} - \mathbf{A}^+\mathbf{A})\boldsymbol{\epsilon}'$ \Comment{forward step}
    \State $L_\theta \gets \left\| \mathcal{\boldsymbol{D}}_{\boldsymbol{\theta}}(\mathbf{x}_t, t) - \mathbf{x} \right\|_1$ \Comment{compute reconstruction loss for $\mathbf{x}$}
    \State $\theta \gets \text{optimizer}(\nabla_\theta L_\theta)$ \Comment{update network parameters $\theta$}
\EndFor
\State \Return $\theta$
\end{algorithmic}
\end{algorithm}

\begin{algorithm}
\caption{SDB Sampling (Euler-Maruyama)}
\label{alg:sdb_sampling}
\begin{algorithmic}[1]
\Require $N, \mathbf{A}, \mathbf{A}^+, \boldsymbol{\Sigma}^{1/2},\alpha_t, \beta_t, \gamma_t, \mathbf{H}_t, \mathbf{F}_t, \mathcal{\boldsymbol{D}}_{\boldsymbol{\theta}}, \mathbf{y}$
\State $t \gets 1$ \Comment{initialize time}
\State $\Delta t \gets 1/N$ \Comment{set timestep}
\State $\boldsymbol{\epsilon}' \sim \mathcal{N}(\mathbf{0}, \mathbf{I}) \in \mathbb{R}^{d}$ \Comment{sample null-space noise}
\State $\hat{\mathbf{x}} \gets \mathbf{A}^+\mathbf{y}$ \Comment{pseudoinverse reconstruction}
\State $\mathbf{x}_t \gets \hat{\mathbf{x}} + \beta_t^{\frac{1}{2}}(\mathbf{I} - \mathbf{A}^+\mathbf{A})\boldsymbol{\epsilon}'$ \Comment{initializer}
\For{$i \in \{1, \ldots, N\}$}
    \State $\boldsymbol{\epsilon} \sim \mathcal{N}(\mathbf{0}, \mathbf{I}) \in \mathbb{R}^{m}$ \Comment{sample range-space noise}
    \State $\boldsymbol{\epsilon}' \sim \mathcal{N}(\mathbf{0}, \mathbf{I}) \in \mathbb{R}^{d}$ \Comment{sample null-space noise}
\State \shortstack[l]{
$\mathbf{x}_{t - \Delta t} \gets \mathbf{x}_t + \Delta t \left[ (f_t\mathbf{I} - 2\mathbf{F}_t)(\mathbf{H}_t\mathcal{\boldsymbol{D}}_{\boldsymbol{\theta}}(\mathbf{x}_t, t) - \mathbf{x}_t) - \mathbf{F}_t\mathbf{x}_t \right] +$ \\
$\Delta t^{1/2} \left[ \left( \frac{\diff \gamma_t}{\diff t}\right)^{\frac{1}{2}} \mathbf{A}^+\boldsymbol{\Sigma}^{\frac{1}{2}}\boldsymbol{\epsilon} + \left( \frac{\diff \beta_t}{\diff t} - 2\beta_t\frac{\diff}{\diff t}\log{\alpha_t} \right)^{\frac{1}{2}}(\mathbf{I} - \mathbf{A}^+\mathbf{A})\boldsymbol{\epsilon}'\right]$
} \Comment{update}
    \State $t \gets t - \Delta t$ \Comment{decrement time}
\EndFor
\State \Return $\mathbf{x}_0$ \Comment{final sample}
\end{algorithmic}
\end{algorithm}

\paragraph{Baselines.}
We compare SDB with both supervised bridge methods and unsupervised plug-and-play diffusion-based baselines. For the former, we include I2SB \citep{liu20232}, IR-SDE \citep{luo2023image}, GOUB \citep{yue2024image}, and DDBM \citep{zhoudenoising}. For the latter, we pick DPS \citep{chung2023diffusion}, $\Pi$GDM \citep{song2023pseudoinverseguided}, and DDNM \citep{wangzero}, all of which rely on the assumption of a noisy linear measurement model. Detailed descriptions of all baselines and other technical details are provided in the Appendix. Unlike these methods, which rely on scalar SDEs, SDB introduces a more general matrix-valued formulation that allows structured control over range and null space components. We summarize the resulting conceptual differences to baseline bridge methods in \cref{tab:comparison}, together with SDB's training and sampling procedures in \cref{alg:sdb_training,alg:sdb_sampling}. Following standard evaluation practice \citep{luo2023image,yue2024image}, we report perceptual scores (FID \citep{heusel2017gans}, LPIPS \citep{zhang2018unreasonable}) and reconstruction metrics (PSNR, SSIM).

\paragraph{Experimental design.}
To isolate the contribution of SDB as a diffusion process, we standardize key implementation details across all methods. Specifically: (i) for both supervised and unsupervised approaches, we train score networks from scratch using the training hyperparameters and architecture of \citet{luo2023image}, with 256 training epochs for supervised methods and 512 for unsupervised ones; (ii) each supervised method learns a mapping between signal samples and their PRs, ensuring that SDB does not benefit from a favorable parameterization, particularly in settings where $m \neq d$; (iii) during evaluation, supervised methods use 100 discretization steps, while unsupervised methods use 200, along with a standard Euler–Maruyama solver and their optimal noise schedule (e.g., VP for DDBM). To ensure consistent evaluation, we reimplement all methods within a unified framework, allowing for the separation of algorithmic advancements from the proposed stochastic process.

\paragraph{Benchmarks.}
We evaluate SDB on four inverse problems with varying measurement system complexities, using original images at a resolution of $256\times 256$. Building on prior work \citep{luo2023image,yue2024image}, we first consider \textbf{inpainting on CelebA-HQ} \citep{karras2018progressive}, where $\mathbf{A}$ is a masking operator and $\boldsymbol{\Sigma} = \mathbf{0}_{m\times m}$ (noiseless). The measurements consist of the masked original images ($m = d$), which are equivalent to the PRs ($\mathbf{A}^+ = \mathbf{I}$). Additionally, we follow previous studies by examining \textbf{superresolution on DIV2K} \citep{Agustsson_2017_CVPR_Workshops,Timofte_2017_CVPR_Workshops}, where $\mathbf{A}$ represents a $4\times$ downsampling operator ($d = 4m$), implemented through average pooling. For this task, $\mathbf{A}^+$ reconstructs the images using nearest-neighbor interpolation.
To evaluate more complex and practical inverse problems, we additionally propose two medical imaging tasks: \textbf{CT reconstruction on the RSNA Intracranial Hemorrhage dataset} \citep{rsna-intracranial-hemorrhage-detection} and \textbf{MRI reconstruction on the Br35H dataset} \citep{merlin9970023542}, both using 2D axial brain scan slices.

For CT reconstruction, the sinogram $\mathbf{y}$ represents line integrals of the object's attenuation coefficient, with the system matrix $\mathbf{A}$ describing detector-specific integrations along X-ray trajectories. We implement $\mathbf{A}$ using its SVD decomposition, $\mathbf{A} = \mathbf{U}\mathbf{D}\mathbf{V}^*$. To make the problem more realistic and introduce domain-specific artifacts, we zero the singular values of $\mathbf{D}$ below a threshold $\tau=3.2$ and use a noisy setting with scalar covariance $\boldsymbol{\Sigma} = \sigma_1^2 \mathbf{I}$, where $\sigma_1^2 = 0.0001$.

For MRI reconstruction, $\mathbf{y}$ consists of undersampled Fourier-domain measurements. The system matrix is modeled as $\mathbf{A} = \mathbf{M} \boldsymbol{\mathcal{F}}$, where $\boldsymbol{\mathcal{F}}$ is the Fourier transform and $\mathbf{M}$ is a masking matrix. We sample $\mathbf{M}$ such that $\lambda_1$ controls the percentage of the lowest frequencies to be kept (a deterministic operation), while $\lambda_2$ specifies the percentage of frequencies to sample from the remaining ones (a stochastic operation). The noisy setting is similarly modeled with a scalar covariance matrix $\boldsymbol{\Sigma} = \sigma_2^2 \mathbf{I}$, where $\sigma_2^2 = 5.0$. We pick $\lambda_1 = 16$ and $\lambda_2 = 30$ as defaults.

\begin{table}[t]
\centering
\caption{Quantitative comparison of SDB with the baselines across four inverse problems. \textbf{Bold} indicates best and \underline{underline} second-best values for each metric.}
\label{tab:main_results}
\setlength{\tabcolsep}{3.5pt}
\renewcommand{\arraystretch}{0.94}
\resizebox{\textwidth}{!}{%
\begin{tabular}{l@{\hskip 3pt}cccc@{\hskip 6pt}cccc@{\hskip 6pt}cccc@{\hskip 6pt}cccc}
\toprule
\multicolumn{1}{c}{} 
& \multicolumn{4}{c}{Inpainting - CelebA-HQ} 
& \multicolumn{4}{c}{Superresolution - DIV2K}
& \multicolumn{4}{c}{CT Reconstruction - RSNA} 
& \multicolumn{4}{c}{MRI Reconstruction - Br35H} \\
\cmidrule(lr){2-5} \cmidrule(lr){6-9} \cmidrule(lr){10-13} \cmidrule(lr){14-17}
Method & FID ↓ & LPIPS ↓ & PSNR ↑ & SSIM ↑
       & FID ↓ & LPIPS ↓ & PSNR ↑ & SSIM ↑
       & FID ↓ & LPIPS ↓ & PSNR ↑ & SSIM ↑
       & FID ↓ & LPIPS ↓ & PSNR ↑ & SSIM ↑ \\
\midrule

\multicolumn{17}{>{\columncolor{gray!20}}l}{\textbf{Unsupervised}} \\
DPS      & 12.3 & 0.213 & 18.59 & 0.684 & 101.2 & 0.295 & 19.04 & 0.492 & 29.25 & 0.160 & 32.870 & 0.709 & 40.73 & 0.187 & 21.336 & 0.549 \\
$\Pi$GDM & 10.2  & 0.146 & 19.02 & 0.762 & 97.09 & 0.311 & 18.84 & 0.530 & 32.29 & 0.165 & 31.272 & 0.757 & 41.14 & 0.211 & 19.830 & 0.665 \\
DDNM     & 10.7  & 0.112 & 18.90 & 0.791 & 99.21 & 0.302 & 17.13 & 0.513 & 30.41 & 0.238 & 27.226 & 0.783 & 42.23 & 0.196 & 18.122 & 0.636 \\

\multicolumn{17}{>{\columncolor{gray!20}}l}{\textbf{Supervised}} \\
I2SB     & 5.56   & 0.047  & 27.41  & 0.889  & \underline{83.73}  & \textbf{0.176}  & 25.21  & 0.686  & 24.81 & 0.107  & 41.886 & 0.924  & 31.54  & 0.065  & 28.750 & 0.849 \\
IR-SDE   & \underline{4.68} & \underline{0.031} & 29.92  & 0.912  & 96.22  & 0.185  & 23.51  & 0.603  & 18.88 & 0.028  & 43.438 & 0.964  & \underline{30.14}  & 0.065  & 28.878 & 0.871 \\
GOUB     & 4.69   & \underline{0.031}  & 29.89  & 0.912  & 98.89  & \underline{0.178}  & 24.39  & 0.649  & 19.90 & 0.024  & 43.878 & 0.967  & 30.63  & \underline{0.058}  & 28.590 & 0.863 \\
DDBM     & 6.03   & 0.047  & 28.15  & 0.906  & 90.16  & 0.233  & 25.79  & \underline{0.720}  & 23.36 & 0.040  & 44.415 & 0.964  & 32.42  & 0.074  & 28.971 & 0.872 \\
\multicolumn{17}{>{\columncolor{gray!20}}l}{\textbf{Ours}} \\
SDB (VP) & 4.90   & \textbf{0.030}  & \underline{30.51}  & 0.914  & 87.08  & 0.228  & \underline{25.91}  & \textbf{0.724}  & 15.43 & \underline{0.019}  & \underline{46.365} & \underline{0.981}  & 32.88  & 0.068  & \underline{29.255} & \underline{0.881} \\
SDB (VE) & 5.97   & 0.042  & \textbf{32.12}  & \textbf{0.944}  & 94.73  & 0.226  & 25.90  & 0.718  & \textbf{14.17} & 0.020  & 46.325 & \underline{0.981}  & 33.90  & 0.083  & 29.098 & 0.876 \\
SDB (SB) & \textbf{4.63}   & \underline{0.031}  & 30.40  & \underline{0.930}  & \textbf{81.56}  & 0.226  & \textbf{26.10}  & \textbf{0.724}  & \underline{15.02} & \textbf{0.018}  & \textbf{46.672} & \textbf{0.982}  & \textbf{29.85}  & \textbf{0.053}  & \textbf{29.812} & \textbf{0.893} \\

\bottomrule
\end{tabular}
}
\end{table}

\subsection{General evaluation}\label{sec:general_eval}

\begin{figure}[t]
    \centering
    \includegraphics[width=1.0\linewidth]{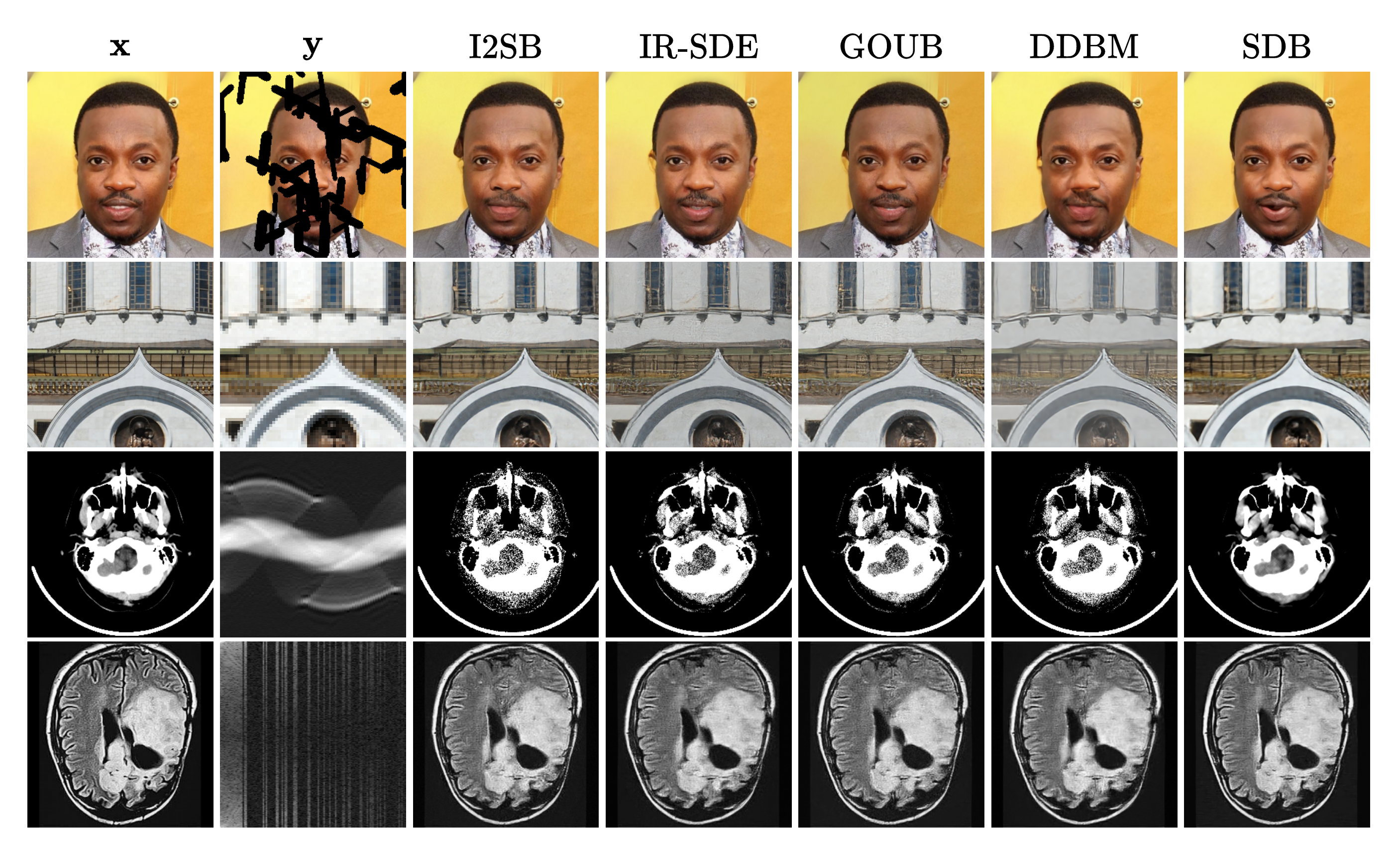}
    \vspace{-1em}
    \caption{Qualitative comparison of SDB (SB) with the best-performing baselines (bridge methods). Rows depict the results for inpainting, superresolution, CT and MRI reconstruction respectively.}
    \label{fig:qual_eval}
\end{figure}

The results summarized in \cref{tab:main_results} show that SDB (SB) outperforms all baseline methods across every metric, with the exception of LPIPS on DIV2K. Notably, SDB (SB) demonstrates a clear advantage over I2SB, which shares the most similar stochastic process, highlighting how SDB (SB) effectively leverages additional information from the measurement system. Overall, it also provides empirical justification for the null space OT plan mentioned in \cref{th:ot_ode}. Moreover, SDB (SB) displays superior stability, as the performance rankings of the baseline bridge methods exhibit more variability across tasks. These advantages are also evident qualitatively (\cref{fig:qual_eval}).

SDB (VE) and SDB (VP) strike a different balance between perceptual and reconstruction metrics compared to SDB (SB). In inpainting, they often outperform all other methods, with a noticeable emphasis on reconstruction quality. This trend is even more pronounced in the superresolution and MRI reconstruction tasks. In CT reconstruction, their performance is nearly on par with SDB (SB). Although they occasionally underperform relative to baseline bridge methods, this is not unexpected given the nature of their stochastic process. Along with IR-SDE, these methods are unique in not treating $p(\mathbf{x}_1)$ as a sum of Dirac deltas, which means they start the reverse process from a noisy sample. This additional noise may make the problem more challenging, but when directly comparing SDB (VE) and SDB (VP) to IR-SDE, the performance improvements become evident.

Finally, under our unified setting, prior bridge methods demonstrate comparable performance with significantly reduced variability compared to what previous works report \citep{luo2023image,yue2024image}. For instance, their PSNR on the MRI reconstruction task falls within the range of $[28.590, 28.971]$, meaning that performance gains on the order of $\approx 0.2$ can be considered significant in this context. Notably, the unsupervised baselines exhibit visibly lower performance compared to bridge methods. While this is expected due to the lack of training with paired data, we note that our setup allocates only twice the training budget to the unconditional models compared to the bridge methods, suggesting that their performance has not yet saturated. We consider improving these models as future work.

\subsection{Evaluation under a misspecified model}\label{sec:misspecified_model}

In practice, while the general form of the measurement system in an inverse problem may be known, the specific parameter values often differ between training and deployment. This creates a generalization challenge, where the model must learn the underlying dynamics rather than relying on shortcuts. For example, in CT reconstruction, a well-trained model should perform stably near the original parameter $\tau$. A larger $\tau'$, which reduces low-frequency information, makes the task more difficult. However, a generalized model should maintain most of its performance as long as $\tau'$ remains reasonably close to $\tau$.

This issue is particularly critical for SDB, where the measurement system's parameters are embedded directly into the coefficients of its stochastic process. While results in \cref{sec:general_eval} indicate performance gains, deploying SDB in practice could be risky if it overfits to specific parameter values. Therefore, the following experiment tests whether SDB can generalize when evaluated under a misspecified model. For this evaluation, we focus on the best-performing methods from \cref{sec:general_eval}, namely the baseline bridge methods and the SB variant of SDB.

\begin{figure}[h]
    \centering
    \includegraphics[width=1.0\linewidth]{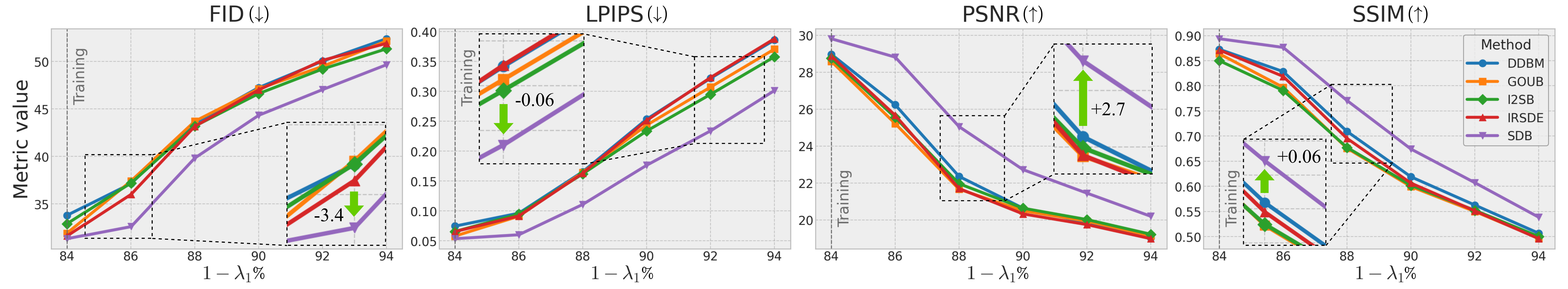}
    \includegraphics[width=1.0\linewidth]{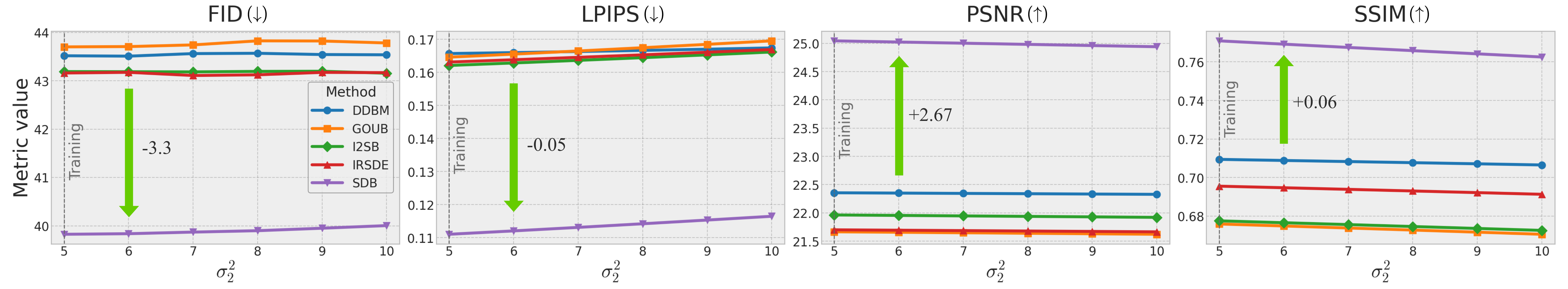}
    \vspace{-1em}
    \caption{Quantitative comparison of SDB (SB) with other bridge methods in a misspecified MRI reconstruction setting. The \textbf{top} row evaluates bridges trained with $\lambda_1 = 16$, $\sigma_2^2 = 5$ on data generated from systems with decreasing $\lambda_1$. The \textbf{bottom} row evaluates performance on data with $\lambda_1 = 14$ and increasing $\sigma_2^2$. Perturbing the original system makes the problem harder in both cases.}
    \label{fig:misspecificed_model}
\vspace{-1em}
\end{figure}

\paragraph{Improved generalization.} We begin with the MRI reconstruction task, where the system response matrix preserves a portion of the original signal by keeping $\lambda_1$ of the frequencies, starting from the lowest. This maintains the general structure of the true image, and as $\lambda_1$ is gradually decreased, more information is lost, making the task increasingly difficult. This scenario also mirrors real-world situations where a detector captures fewer measurements.

In what follows, we evaluate the checkpoints of bridge methods trained under a default setting ($\lambda_1=16,\lambda_2=30,\sigma_2^2=5.0$) on measurements generated from a system with modified parameters. \Cref{fig:misspecificed_model} (top) shows the performance of the methods as $\lambda_1$ is gradually decreased. Notably, the performance gap between SDB and the baseline methods widens as early as $\lambda_1 = 14$, and this advantage is sustained even at lower values. SDB continues to better utilize the available measurement information, indicating much greater robustness to reductions in $\lambda_1$.

To make the setting even more challenging, we set $\lambda_1 = 14$ and gradually increase the measurement noise variance $\sigma_2^2$ to up to twice its training-time value, simulating a more realistic scenario. \Cref{fig:misspecificed_model} (bottom) shows that, even under this perturbation, SDB maintains a significant performance advantage over the baseline methods. Unsurprisingly, all methods exhibit greater robustness to changes in $\sigma_2^2$, which is expected due to their denoising nature.

An analogous analysis for the CT reconstruction task is provided in the Appendix, yielding similar observations. Both experiments, closely aligned with the practical applications of inverse problem solvers, underscore the generalization capabilities of SDB across a broad range of system perturbations, positioning it as a promising solution for real-world scenarios.

\paragraph{Extended evaluation.}
Additional experimental results are presented in the Appendix. These include evaluations on more challenging misspecified models, analyses of the influence of ill-conditioned system operators, and detailed assessments of computational complexity and runtime. We also include a series of ablation studies. Finally, we outline a conceptual extension of SDB to nonlinear measurement systems and demonstrate its empirical advantages over baseline bridge methods in a small-scale study. Extending SDB to a broader class of problems remains an interesting direction for future work.

\section{Discussion and limitations}

SDB offers a principled framework for constructing measurement-system-specific diffusion bridges tailored to inverse problems under a linear Gaussian model. By explicitly incorporating information about the measurement system into the generative process, it achieves improved reconstruction quality and exhibits strong generalization under system perturbations. However, the method also comes with natural limitations. Notably, extending SDB to nonlinear measurement systems remains an open challenge. We provide only a preliminary, proof-of-concept treatment based on local linearization, which is limited to differentiable systems. In addition, our work adopts simple variance schedules for both the range and null space components; the interplay between these two processes warrants further theoretical and empirical investigation to identify optimal scheduling strategies. From a real-world application perspective, tasks such as CT and MRI reconstruction are typically performed in three dimensions—evaluating our method in such settings remains an open direction for future work. Despite these limitations, we view SDB as a valuable contribution to the growing literature on diffusion-based inverse problem solvers and a promising foundation for future research.

\newpage
\section*{Acknowledgments}

This work was financially supported by the INFOSTRATEG-I/0022/2021-00 grant funded by the Polish National Centre for Research and Development (NCBiR), the SONATA BIS grant 2019/34/E/ST6/00052 funded by the Polish National Science Centre (NCN), and the NIH grant 5R01HL159183-03.

The computational resources for this work were provided by the Laboratory of Bioinformatics and Computational Genomics and the High Performance Computing Center of the Faculty of Mathematics and Information Science, Warsaw University of Technology. We also gratefully acknowledge Poland's High-performance Infrastructure PLGrid ACC Cyfronet AGH for providing computer facilities and support within computational grant no. PLG/2025/018330.

\bibliographystyle{plainnat}
\bibliography{main}


\newpage
\section*{NeurIPS Paper Checklist}

\begin{enumerate}

\item {\bf Claims}
    \item[] Question: Do the main claims made in the abstract and introduction accurately reflect the paper's contributions and scope?
    \item[] Answer: \answerYes{} 
    \item[] Justification: The claims are supported with experimental and theoretical results.
    \item[] Guidelines:
    \begin{itemize}
        \item The answer NA means that the abstract and introduction do not include the claims made in the paper.
        \item The abstract and/or introduction should clearly state the claims made, including the contributions made in the paper and important assumptions and limitations. A No or NA answer to this question will not be perceived well by the reviewers. 
        \item The claims made should match theoretical and experimental results, and reflect how much the results can be expected to generalize to other settings. 
        \item It is fine to include aspirational goals as motivation as long as it is clear that these goals are not attained by the paper. 
    \end{itemize}

\item {\bf Limitations}
    \item[] Question: Does the paper discuss the limitations of the work performed by the authors?
    \item[] Answer: \answerYes{} 
    \item[] Justification: The limitations are described in the discussion section.
    \item[] Guidelines:
    \begin{itemize}
        \item The answer NA means that the paper has no limitation while the answer No means that the paper has limitations, but those are not discussed in the paper. 
        \item The authors are encouraged to create a separate "Limitations" section in their paper.
        \item The paper should point out any strong assumptions and how robust the results are to violations of these assumptions (e.g., independence assumptions, noiseless settings, model well-specification, asymptotic approximations only holding locally). The authors should reflect on how these assumptions might be violated in practice and what the implications would be.
        \item The authors should reflect on the scope of the claims made, e.g., if the approach was only tested on a few datasets or with a few runs. In general, empirical results often depend on implicit assumptions, which should be articulated.
        \item The authors should reflect on the factors that influence the performance of the approach. For example, a facial recognition algorithm may perform poorly when image resolution is low or images are taken in low lighting. Or a speech-to-text system might not be used reliably to provide closed captions for online lectures because it fails to handle technical jargon.
        \item The authors should discuss the computational efficiency of the proposed algorithms and how they scale with dataset size.
        \item If applicable, the authors should discuss possible limitations of their approach to address problems of privacy and fairness.
        \item While the authors might fear that complete honesty about limitations might be used by reviewers as grounds for rejection, a worse outcome might be that reviewers discover limitations that aren't acknowledged in the paper. The authors should use their best judgment and recognize that individual actions in favor of transparency play an important role in developing norms that preserve the integrity of the community. Reviewers will be specifically instructed to not penalize honesty concerning limitations.
    \end{itemize}

\item {\bf Theory assumptions and proofs}
    \item[] Question: For each theoretical result, does the paper provide the full set of assumptions and a complete (and correct) proof?
    \item[] Answer: \answerYes{} 
    \item[] Justification: The theory and assumptions are clearly stated and proven in the methodological section and the Appendix.
    \item[] Guidelines:
    \begin{itemize}
        \item The answer NA means that the paper does not include theoretical results. 
        \item All the theorems, formulas, and proofs in the paper should be numbered and cross-referenced.
        \item All assumptions should be clearly stated or referenced in the statement of any theorems.
        \item The proofs can either appear in the main paper or the supplemental material, but if they appear in the supplemental material, the authors are encouraged to provide a short proof sketch to provide intuition. 
        \item Inversely, any informal proof provided in the core of the paper should be complemented by formal proofs provided in appendix or supplemental material.
        \item Theorems and Lemmas that the proof relies upon should be properly referenced. 
    \end{itemize}

    \item {\bf Experimental result reproducibility}
    \item[] Question: Does the paper fully disclose all the information needed to reproduce the main experimental results of the paper to the extent that it affects the main claims and/or conclusions of the paper (regardless of whether the code and data are provided or not)?
    \item[] Answer: \answerYes{} 
    \item[] Justification: All experimental methods are mentioned in the main paper and the Appendix.
    \item[] Guidelines:
    \begin{itemize}
        \item The answer NA means that the paper does not include experiments.
        \item If the paper includes experiments, a No answer to this question will not be perceived well by the reviewers: Making the paper reproducible is important, regardless of whether the code and data are provided or not.
        \item If the contribution is a dataset and/or model, the authors should describe the steps taken to make their results reproducible or verifiable. 
        \item Depending on the contribution, reproducibility can be accomplished in various ways. For example, if the contribution is a novel architecture, describing the architecture fully might suffice, or if the contribution is a specific model and empirical evaluation, it may be necessary to either make it possible for others to replicate the model with the same dataset, or provide access to the model. In general. releasing code and data is often one good way to accomplish this, but reproducibility can also be provided via detailed instructions for how to replicate the results, access to a hosted model (e.g., in the case of a large language model), releasing of a model checkpoint, or other means that are appropriate to the research performed.
        \item While NeurIPS does not require releasing code, the conference does require all submissions to provide some reasonable avenue for reproducibility, which may depend on the nature of the contribution. For example
        \begin{enumerate}
            \item If the contribution is primarily a new algorithm, the paper should make it clear how to reproduce that algorithm.
            \item If the contribution is primarily a new model architecture, the paper should describe the architecture clearly and fully.
            \item If the contribution is a new model (e.g., a large language model), then there should either be a way to access this model for reproducing the results or a way to reproduce the model (e.g., with an open-source dataset or instructions for how to construct the dataset).
            \item We recognize that reproducibility may be tricky in some cases, in which case authors are welcome to describe the particular way they provide for reproducibility. In the case of closed-source models, it may be that access to the model is limited in some way (e.g., to registered users), but it should be possible for other researchers to have some path to reproducing or verifying the results.
        \end{enumerate}
    \end{itemize}

\item {\bf Open access to data and code}
    \item[] Question: Does the paper provide open access to the data and code, with sufficient instructions to faithfully reproduce the main experimental results, as described in supplemental material?
    \item[] Answer: \answerYes{} 
    \item[] Justification: The data and code are provided as open-source.
    \item[] Guidelines:
    \begin{itemize}
        \item The answer NA means that paper does not include experiments requiring code.
        \item Please see the NeurIPS code and data submission guidelines (\url{https://nips.cc/public/guides/CodeSubmissionPolicy}) for more details.
        \item While we encourage the release of code and data, we understand that this might not be possible, so “No” is an acceptable answer. Papers cannot be rejected simply for not including code, unless this is central to the contribution (e.g., for a new open-source benchmark).
        \item The instructions should contain the exact command and environment needed to run to reproduce the results. See the NeurIPS code and data submission guidelines (\url{https://nips.cc/public/guides/CodeSubmissionPolicy}) for more details.
        \item The authors should provide instructions on data access and preparation, including how to access the raw data, preprocessed data, intermediate data, and generated data, etc.
        \item The authors should provide scripts to reproduce all experimental results for the new proposed method and baselines. If only a subset of experiments are reproducible, they should state which ones are omitted from the script and why.
        \item At submission time, to preserve anonymity, the authors should release anonymized versions (if applicable).
        \item Providing as much information as possible in supplemental material (appended to the paper) is recommended, but including URLs to data and code is permitted.
    \end{itemize}

\item {\bf Experimental setting/details}
    \item[] Question: Does the paper specify all the training and test details (e.g., data splits, hyperparameters, how they were chosen, type of optimizer, etc.) necessary to understand the results?
    \item[] Answer: \answerYes{} 
    \item[] Justification: All training and evaluation details are provided in the main paper and the Appendix.
    \item[] Guidelines:
    \begin{itemize}
        \item The answer NA means that the paper does not include experiments.
        \item The experimental setting should be presented in the core of the paper to a level of detail that is necessary to appreciate the results and make sense of them.
        \item The full details can be provided either with the code, in appendix, or as supplemental material.
    \end{itemize}

\item {\bf Experiment statistical significance}
    \item[] Question: Does the paper report error bars suitably and correctly defined or other appropriate information about the statistical significance of the experiments?
    \item[] Answer: \answerNo{} 
    \item[] Justification: Repeating the experimental evaluation many times entails a significant computational burden. Care was taken to ensure that the performance of each method is reported in a fair and representative manner.
    \item[] Guidelines:
    \begin{itemize}
        \item The answer NA means that the paper does not include experiments.
        \item The authors should answer "Yes" if the results are accompanied by error bars, confidence intervals, or statistical significance tests, at least for the experiments that support the main claims of the paper.
        \item The factors of variability that the error bars are capturing should be clearly stated (for example, train/test split, initialization, random drawing of some parameter, or overall run with given experimental conditions).
        \item The method for calculating the error bars should be explained (closed form formula, call to a library function, bootstrap, etc.)
        \item The assumptions made should be given (e.g., Normally distributed errors).
        \item It should be clear whether the error bar is the standard deviation or the standard error of the mean.
        \item It is OK to report 1-sigma error bars, but one should state it. The authors should preferably report a 2-sigma error bar than state that they have a 96\% CI, if the hypothesis of Normality of errors is not verified.
        \item For asymmetric distributions, the authors should be careful not to show in tables or figures symmetric error bars that would yield results that are out of range (e.g. negative error rates).
        \item If error bars are reported in tables or plots, The authors should explain in the text how they were calculated and reference the corresponding figures or tables in the text.
    \end{itemize}

\item {\bf Experiments compute resources}
    \item[] Question: For each experiment, does the paper provide sufficient information on the computer resources (type of compute workers, memory, time of execution) needed to reproduce the experiments?
    \item[] Answer: \answerYes{} 
    \item[] Justification: Computational details are included in the Appendix.
    \item[] Guidelines:
    \begin{itemize}
        \item The answer NA means that the paper does not include experiments.
        \item The paper should indicate the type of compute workers CPU or GPU, internal cluster, or cloud provider, including relevant memory and storage.
        \item The paper should provide the amount of compute required for each of the individual experimental runs as well as estimate the total compute. 
        \item The paper should disclose whether the full research project required more compute than the experiments reported in the paper (e.g., preliminary or failed experiments that didn't make it into the paper). 
    \end{itemize}
    
\item {\bf Code of ethics}
    \item[] Question: Does the research conducted in the paper conform, in every respect, with the NeurIPS Code of Ethics \url{https://neurips.cc/public/EthicsGuidelines}?
    \item[] Answer: \answerYes{} 
    \item[] Justification: All conducted research conforms to the Code of Ethics.
    \item[] Guidelines:
    \begin{itemize}
        \item The answer NA means that the authors have not reviewed the NeurIPS Code of Ethics.
        \item If the authors answer No, they should explain the special circumstances that require a deviation from the Code of Ethics.
        \item The authors should make sure to preserve anonymity (e.g., if there is a special consideration due to laws or regulations in their jurisdiction).
    \end{itemize}

\item {\bf Broader impacts}
    \item[] Question: Does the paper discuss both potential positive societal impacts and negative societal impacts of the work performed?
    \item[] Answer: \answerYes{} 
    \item[] Justification: Impacts are discussed in the introduction and the Appendix.
    \item[] Guidelines:
    \begin{itemize}
        \item The answer NA means that there is no societal impact of the work performed.
        \item If the authors answer NA or No, they should explain why their work has no societal impact or why the paper does not address societal impact.
        \item Examples of negative societal impacts include potential malicious or unintended uses (e.g., disinformation, generating fake profiles, surveillance), fairness considerations (e.g., deployment of technologies that could make decisions that unfairly impact specific groups), privacy considerations, and security considerations.
        \item The conference expects that many papers will be foundational research and not tied to particular applications, let alone deployments. However, if there is a direct path to any negative applications, the authors should point it out. For example, it is legitimate to point out that an improvement in the quality of generative models could be used to generate deepfakes for disinformation. On the other hand, it is not needed to point out that a generic algorithm for optimizing neural networks could enable people to train models that generate Deepfakes faster.
        \item The authors should consider possible harms that could arise when the technology is being used as intended and functioning correctly, harms that could arise when the technology is being used as intended but gives incorrect results, and harms following from (intentional or unintentional) misuse of the technology.
        \item If there are negative societal impacts, the authors could also discuss possible mitigation strategies (e.g., gated release of models, providing defenses in addition to attacks, mechanisms for monitoring misuse, mechanisms to monitor how a system learns from feedback over time, improving the efficiency and accessibility of ML).
    \end{itemize}
    
\item {\bf Safeguards}
    \item[] Question: Does the paper describe safeguards that have been put in place for responsible release of data or models that have a high risk for misuse (e.g., pretrained language models, image generators, or scraped datasets)?
    \item[] Answer: \answerNo{} 
    \item[] Justification: The method was applied to open-source data and poses a low risk of misuse.
    \item[] Guidelines:
    \begin{itemize}
        \item The answer NA means that the paper poses no such risks.
        \item Released models that have a high risk for misuse or dual-use should be released with necessary safeguards to allow for controlled use of the model, for example by requiring that users adhere to usage guidelines or restrictions to access the model or implementing safety filters. 
        \item Datasets that have been scraped from the Internet could pose safety risks. The authors should describe how they avoided releasing unsafe images.
        \item We recognize that providing effective safeguards is challenging, and many papers do not require this, but we encourage authors to take this into account and make a best faith effort.
    \end{itemize}

\item {\bf Licenses for existing assets}
    \item[] Question: Are the creators or original owners of assets (e.g., code, data, models), used in the paper, properly credited and are the license and terms of use explicitly mentioned and properly respected?
    \item[] Answer: \answerYes{} 
    \item[] Justification: Datasets and code are credited.
    \item[] Guidelines:
    \begin{itemize}
        \item The answer NA means that the paper does not use existing assets.
        \item The authors should cite the original paper that produced the code package or dataset.
        \item The authors should state which version of the asset is used and, if possible, include a URL.
        \item The name of the license (e.g., CC-BY 4.0) should be included for each asset.
        \item For scraped data from a particular source (e.g., website), the copyright and terms of service of that source should be provided.
        \item If assets are released, the license, copyright information, and terms of use in the package should be provided. For popular datasets, \url{paperswithcode.com/datasets} has curated licenses for some datasets. Their licensing guide can help determine the license of a dataset.
        \item For existing datasets that are re-packaged, both the original license and the license of the derived asset (if it has changed) should be provided.
        \item If this information is not available online, the authors are encouraged to reach out to the asset's creators.
    \end{itemize}

\item {\bf New assets}
    \item[] Question: Are new assets introduced in the paper well documented and is the documentation provided alongside the assets?
    \item[] Answer: \answerNA{} 
    \item[] Justification: No new assets were provided.
    \item[] Guidelines:
    \begin{itemize}
        \item The answer NA means that the paper does not release new assets.
        \item Researchers should communicate the details of the dataset/code/model as part of their submissions via structured templates. This includes details about training, license, limitations, etc. 
        \item The paper should discuss whether and how consent was obtained from people whose asset is used.
        \item At submission time, remember to anonymize your assets (if applicable). You can either create an anonymized URL or include an anonymized zip file.
    \end{itemize}

\item {\bf Crowdsourcing and research with human subjects}
    \item[] Question: For crowdsourcing experiments and research with human subjects, does the paper include the full text of instructions given to participants and screenshots, if applicable, as well as details about compensation (if any)? 
    \item[] Answer: \answerNA{} 
    \item[] Justification: No human subjects were involved.
    \item[] Guidelines:
    \begin{itemize}
        \item The answer NA means that the paper does not involve crowdsourcing nor research with human subjects.
        \item Including this information in the supplemental material is fine, but if the main contribution of the paper involves human subjects, then as much detail as possible should be included in the main paper. 
        \item According to the NeurIPS Code of Ethics, workers involved in data collection, curation, or other labor should be paid at least the minimum wage in the country of the data collector. 
    \end{itemize}

\item {\bf Institutional review board (IRB) approvals or equivalent for research with human subjects}
    \item[] Question: Does the paper describe potential risks incurred by study participants, whether such risks were disclosed to the subjects, and whether Institutional Review Board (IRB) approvals (or an equivalent approval/review based on the requirements of your country or institution) were obtained?
    \item[] Answer: \answerNA{} 
    \item[] Justification: Not related to the conducted research.
    \item[] Guidelines:
    \begin{itemize}
        \item The answer NA means that the paper does not involve crowdsourcing nor research with human subjects.
        \item Depending on the country in which research is conducted, IRB approval (or equivalent) may be required for any human subjects research. If you obtained IRB approval, you should clearly state this in the paper. 
        \item We recognize that the procedures for this may vary significantly between institutions and locations, and we expect authors to adhere to the NeurIPS Code of Ethics and the guidelines for their institution. 
        \item For initial submissions, do not include any information that would break anonymity (if applicable), such as the institution conducting the review.
    \end{itemize}

\item {\bf Declaration of LLM usage}
    \item[] Question: Does the paper describe the usage of LLMs if it is an important, original, or non-standard component of the core methods in this research? Note that if the LLM is used only for writing, editing, or formatting purposes and does not impact the core methodology, scientific rigorousness, or originality of the research, declaration is not required.
    \item[] Answer: \answerNA{} 
    \item[] Justification: LLMs are only used for editing and formatting purposes.
    \item[] Guidelines:
    \begin{itemize}
        \item The answer NA means that the core method development in this research does not involve LLMs as any important, original, or non-standard components.
        \item Please refer to our LLM policy (\url{https://neurips.cc/Conferences/2025/LLM}) for what should or should not be described.
    \end{itemize}

\end{enumerate}

\newpage

\appendix
\section*{Appendix for "System-Embedded Diffusion Bridge Models"}
\startcontents[sections]
\printcontents[sections]{l}{1}{\setcounter{tocdepth}{2}}

\section{Theoretical results, proofs and derivations}

\subsection{Derivation of drift and diffusion coefficients for SDB}

We begin with deriving the drift and diffusion coefficients $\mathbf{F}_t, \mathbf{G}_t$ of SDB (\cref{eq:sdb_drift,eq:sdb_diffusion}) by using \cref{th:sde_to_cascade} with the mean and covariance matrices $\mathbf{H}_t, \boldsymbol{\Sigma}_t$ (\cref{eq:sdb_cascade_linear,eq:sdb_cascade_covar}) of its forward process. Recall that $\mathbf{H}_t=\mathbf{A}^+\mathbf{A} + \alpha_t(\mathbf{I} - \mathbf{A}^+\mathbf{A}),
\boldsymbol{\Sigma}_t=\gamma_t\mathbf{A}^+\boldsymbol{\Sigma}{\mathbf{A}^{+}}^\top + \beta_t(\mathbf{I} - \mathbf{A}^+\mathbf{A})$. Following \cref{th:sde_to_cascade}, we obtain
\begin{subequations}
\begin{align}
\mathbf{A}^+\mathbf{A}\mathbf{F}_t=&\frac{\diff}{\diff t}\mathbf{A}^+\mathbf{A}\log{\mathbf{H}_t} = \frac{\diff}{\diff t}\mathbf{A}^+\mathbf{A} = \mathbf{0},\\
(\mathbf{I} - \mathbf{A^+}\mathbf{A})\mathbf{F}_t=&\frac{\diff}{\diff t}\log{\alpha_t}(\mathbf{I} - \mathbf{A^+}\mathbf{A}),\\
\frac{\diff}{\diff t}\mathbf{A}^+\mathbf{A} \boldsymbol{\Sigma}_t=&\frac{\diff\gamma_t}{\diff t}\mathbf{A}^+\boldsymbol{\Sigma}\mathbf{A}^{+^\top},\\
\frac{\diff}{\diff t}(\mathbf{I - \mathbf{A}^+\mathbf{A})}\boldsymbol{\Sigma}_t =& \frac{\diff \beta_t}{\diff t}(\mathbf{I} - \mathbf{A}^+\mathbf{A}).
\end{align}
\end{subequations}
Moreover, note that $\mathbf{F}_t^\top=\mathbf{F}_t,\boldsymbol{\Sigma}_t^\top=\boldsymbol{\Sigma}_t$ due to the symmetry of the range and null space projections, and $\mathbf{F}_t\boldsymbol{\Sigma}_t=\boldsymbol{\Sigma}_t\mathbf{F}_t$ since $\mathbf{F}_t$ only affects the null space and $(\mathbf{I} - \mathbf{A}^+\mathbf{A})^2=(\mathbf{I} - \mathbf{A}^+\mathbf{A})$ (idempotent). Following \cref{th:sde_to_cascade},
\begin{subequations}
    \begin{align}
        \mathbf{F}_t=&\frac{\diff}{\diff t}\log{\alpha}_t(\mathbf{I} - \mathbf{A}^+\mathbf{A}),\\
        \mathbf{G}_t\mathbf{G}_t=&\frac{\diff \boldsymbol{\Sigma}_t}{\diff t} - \mathbf{F}_t\boldsymbol{\Sigma}_t-\boldsymbol{\Sigma}_t\mathbf{F}_t^\top\\=&
        \frac{\diff \boldsymbol{\Sigma}_t}{\diff t} - 2\mathbf{F}_t\boldsymbol{\Sigma}_t\\
        =&\frac{\diff \gamma_t}{\diff t}\mathbf{A}^+\boldsymbol{\Sigma}\mathbf{A}^{+^\top} + \left(\frac{\diff \beta_t}{\diff t} - 2\beta_t \frac{\diff}{\diff t}\log{\alpha_t}\right)(\mathbf{I} - \mathbf{A}^+\mathbf{A}).
    \end{align}
\end{subequations}

\subsection{Proof of \cref{th:ot_ode}}
We begin by recalling two propositions of \citet{liu20232} using our notation, which will serve as the basis of the proof.

\begin{proposition}\textbf{(Analytic posterior given boundary pair) \citep{liu20232}}
The posterior of \cref{eq:linear_forward_sde} given some boundary pair $(\mathbf{x}_0, \mathbf{x}_1)$ admits an analytic form:
\begin{equation}
    p(\mathbf{x}_t \mid \mathbf{x}_0, \mathbf{x}_1)=\boldsymbol{\mathcal{N}}(\boldsymbol{\mu}_t(\mathbf{x}_0, \mathbf{x}_1), \boldsymbol{\Sigma}_t),
    \label{eq:i2sb_posterior}
\end{equation}
where $\boldsymbol{\mu}_t=\frac{\bar{\sigma}_t^2}{\bar{\sigma}_t^2 + \sigma_t^2}\mathbf{x}_0 + \frac{\sigma_t^2}{\bar{\sigma}_t^2 + \sigma_t^2}\mathbf{x}_1,\boldsymbol{\Sigma}_t=\frac{\sigma_t^2\bar{\sigma}_t^2}{\bar{\sigma}_t^2 + \sigma_t^2}\mathbf{I}$ and $\sigma_t^2=\int_0^tg^2(\tau)\diff\tau, \bar{\sigma}_t^2=\int_t^1=g^2(\tau)\diff\tau$ with $\boldsymbol{G}_t=g(t)\mathbf{I}$.
\label{prop:i2sb_posterior}
\end{proposition}

\begin{proposition}\textbf{(Optimal Transport ODE; OT-ODE) \citep{liu20232}}
When $g^2(t)\to 0$, the SDE between ($\mathbf{x}_0, \mathbf{x}_1$) reduces to an ODE:
\begin{equation}
    \diff \mathbf{x}_t=\mathbf{v}_t(\mathbf{x}_t\mid \mathbf{x}_0)\diff t,
\end{equation}
where $\mathbf{v}_t(\mathbf{x}_t\mid \mathbf{x}_0)=\frac{g^2(t)}{\sigma_t^2}(\mathbf{x}_t - \mathbf{x}_0)$ whose solution is the posterior mean of \cref{eq:i2sb_posterior}.
\label{prop:i2sb_ot_ode}
\end{proposition}

Consider the null space part of $\mathbf{x}_t$ given by mean and covariance matrices from \cref{eq:sdb_cascade_linear,eq:sdb_cascade_covar} with $\alpha_t=\frac{\bar{\sigma}_t^2}{\bar{\sigma}_t^2 + \sigma_t^2},\beta_t=\frac{\sigma_t^2\bar{\sigma}_t^2}{\bar{\sigma}_t^2 + \sigma_t^2}$ for $\sigma_t^2=\int_0^t g^2(\tau)d\tau, \bar{\sigma}_t^2=\int_t^1 g^2(\tau)d\tau$, where $g(t)$ is the null space diffusion coefficient given by \cref{eq:sdb_diffusion}, \ie, $g(t)=\left(\frac{\diff \beta_t}{\diff t} - 2\beta_t \frac{\diff}{\diff t}\log{\alpha_t}\right)^{\frac{1}{2}}$:
\begin{equation}
(\mathbf{I} - \mathbf{A}^+ \mathbf{A})\mathbf{x}_t=\frac{\bar{\sigma}_t^2}{\bar{\sigma}_t^2 + \sigma_t^2}(\mathbf{I} - \mathbf{A}^+ \mathbf{A})\mathbf{x}_0 + \left(\frac{\sigma_t^2\bar{\sigma}_t^2}{\bar{\sigma}_t^2 + \sigma_t^2}\right)^{\frac{1}{2}}(\mathbf{I} - \mathbf{A}^+ \mathbf{A})\boldsymbol{\epsilon}
\end{equation}
for $\boldsymbol{\epsilon}\sim\boldsymbol{\mathcal{N}}(\mathbf{0}_{d}, \mathbf{I}_{d\times d}).$ Note that the null space part of $\mathbf{x}_1$, being the PR of $\mathbf{x}_0$, is zeroed out, \ie, $(\mathbf{I} - \mathbf{A}^+\mathbf{A})\mathbf{x}_1=(\mathbf{I} - \mathbf{A}^+\mathbf{A})\mathbf{0}_{d\times d}.$ This allows us to artificially rewrite $\mathbf{x}_t$ as
\begin{equation}
(\mathbf{I} - \mathbf{A}^+ \mathbf{A})\mathbf{x}_t=(\mathbf{I} - \mathbf{A}^+ \mathbf{A})\left(\frac{\bar{\sigma}_t^2}{\bar{\sigma}_t^2 + \sigma_t^2}\mathbf{x}_0 + \frac{\sigma_t^2}{\bar{\sigma}_t^2 + \sigma_t^2}\mathbf{0}_{d\times d}\right) + \left(\frac{\sigma_t^2\bar{\sigma}_t^2}{\bar{\sigma}_t^2 + \sigma_t^2}\right)^{\frac{1}{2}}(\mathbf{I} - \mathbf{A}^+ \mathbf{A})\boldsymbol{\epsilon}.
\end{equation}
Equivalently, $(\mathbf{I} - \mathbf{A}^+\mathbf{A})\mathbf{x}_t\sim\boldsymbol{\mathcal{N}}((\mathbf{I} - \mathbf{A}^+ \mathbf{A})\left(\frac{\bar{\sigma}_t^2}{\bar{\sigma}_t^2 + \sigma_t^2}\mathbf{x}_0 + \frac{\sigma_t^2}{\bar{\sigma}_t^2 + \sigma_t^2}\mathbf{x}_1\right),\frac{\sigma_t^2\bar{\sigma}_t^2}{\bar{\sigma}_t^2 + \sigma_t^2}(\mathbf{I} - \mathbf{A}^+ \mathbf{A}))$. Hence, the posterior mean of SDB (SB) takes the form stated in \cref{prop:i2sb_posterior} and \cref{prop:i2sb_ot_ode} can be applied directly to the null space part.

As a final remark, we note that, up to this point, the definition of $g(t)$ is interdependent with that of $\alpha_t$ and $\beta_t$. It is not immediately clear that defining $\alpha_t, \beta_t$ as in \cref{th:ot_ode} fulfills $g^2(t)=\frac{\diff \beta_t}{\diff t} - 2\beta_t \frac{\diff}{\diff t}\log{\alpha_t}$. We now show that this property is indeed satisfied.

Denote by $C=\sigma_t^2 + \bar{\sigma}_t^2$ and observe that $\beta_t=\sigma_t^2\alpha_t,\bar{\sigma}_t^2=C - \sigma_t^2,\frac{\diff\sigma_t^2}{\diff t}=g^2(t)$. Then,
\begin{subequations}
    \begin{align}
        \frac{\diff}{\diff t}\log{\alpha_t}=&\frac{\diff}{\diff t} \log{\frac{\bar{\sigma}_t^2}{\bar{\sigma}_t^2 + \sigma_t^2}}\\
        =&\frac{\diff}{\diff t}\left( \log{\bar{\sigma}_t^2} - \log{C} \right)\\
        =&\frac{g^2(t)}{\bar{\sigma}_t^2},\\
        \frac{\diff \beta_t}{\diff t} =& \frac{\diff}{\diff t} (\frac{\sigma_t^2\bar{\sigma}_t^2}{\bar{\sigma}_t^2+ \sigma_t^2})\\
        =& \frac{\diff}{\diff t} \left( \frac{\sigma_t^2(C - \sigma_t^2)}{C}\right)\\
        =& \frac{1}{C}\frac{\diff}{\diff t}\left( C\sigma_t^2 - \sigma_t^4\right)\\
        =& g^2(t) - \frac{2}{C}\sigma_t^2g^2(t)\\
        =& g^2(t)(1 - \frac{2}{C}\sigma_t^2).
    \end{align}
\end{subequations}
By substituting these into the definition of $g^2(t)$, we obtain
\begin{subequations}
    \begin{align}
        \frac{\diff \beta_t}{\diff t} - 2\beta_t \frac{\diff}{\diff t}\log{\alpha_t}=&\frac{\diff}{\diff t}(\sigma_t^2 \alpha_t) - 2\sigma_t^2\alpha_t\frac{\diff}{\diff t}\log{\alpha_t}\\
        =&\alpha_t\frac{\diff \sigma_t^2}{\diff t} + \sigma_t^2\frac{\diff \alpha_t}{\diff t} - 2\sigma_t^2\alpha_t\left( \frac{1}{\alpha_t}\frac{\diff\alpha_t}{\diff t}\right)\\
        =&\alpha_tg^2(t) - \sigma_t^2\frac{\diff\alpha_t}{\diff t}\\
        =&\left( 1 - \frac{\sigma_t^2}{C}\right)g^2(t) + \frac{\sigma_t^2}{C}g^2(t)\\
        =&g^2(t),
    \end{align}
\end{subequations}
which completes the proof.

\hfill $\square$

\subsection{Proof of \cref{th:principled_posterior_sampler}}

Many diffusion bridge methods, such as I2SB, GOUB, or DDBM, rely on non-Markovian stochastic processes, where $\mathbf{x}_t$ depends on both $\mathbf{x}_0$ and $\mathbf{x}_1$. In contrast, SDB uses a Markovian forward diffusion process, for which $p(\mathbf{x}_t|\mathbf{x}_0)$ is well-defined. More specifically, SDB is a special case of score-based generative models, using the general formulation with matrix-valued drift and diffusion coefficients described in Appendix A of~\citep{songscore}. We show below that SDB is a principled posterior sampler of $p(\mathbf{x}|\mathbf{y})$ with a proof of \cref{th:principled_posterior_sampler} based on a series of lemmas.



We start by considering the most general practical case with a positive semidefinite covariance matrix $\boldsymbol{\Sigma}$ in the forward model $p(\mathbf{y}' \mid \mathbf{x}) = \mathcal{N}(\mathbf{A}'\mathbf{x}, \boldsymbol{\Sigma})$. When $\boldsymbol{\Sigma}$ is positive definite, we can perform whitening by scaling the measurements and the system matrix as $\mathbf{y} = \boldsymbol{\Sigma}^{-\frac{1}{2}}\mathbf{y}'$ and $\mathbf{A} = \boldsymbol{\Sigma}^{-\frac{1}{2}}\mathbf{A}'$, yielding the equivalent isotropic model $p(\mathbf{y} \mid \mathbf{x}) = \mathcal{N}(\mathbf{A}\mathbf{x}, \mathbf{I})$. In the singular case (i.e., $\boldsymbol{\Sigma}$ is positive semidefinite but not positive definite), $\boldsymbol{\Sigma}^{-\frac{1}{2}}$ can be replaced with $(\boldsymbol{\Sigma}^+)^{\frac{1}{2}}$ to achieve whitening on the range space of $\boldsymbol{\Sigma}$. From this point onward, we drop primes for notational simplicity and proceed in the whitened coordinates. Hence, without loss of generality, we assume an isotropic noise model, which is equivalent to an initial whitening of both the measurements and the system response matrix.

\begin{lemma}\label{lemma:pinv_suff}
The pseudoinverse reconstruction $\mathbf{A}^{+}\mathbf{y}$ is a sufficient statistic for estimation of $\mathbf{x}$ given $\mathbf{y}$. That is,
\[
p(\mathbf{x} | \mathbf{y}) = p(\mathbf{x} | \mathbf{A}^{+} \mathbf{y}).
\]
\end{lemma}

This holds because (i) the range space component of the measurements can be exactly recovered by re-applying $\mathbf{A}$ to $\mathbf{A}^{+} \mathbf{y}$, and (ii) the null space component contains no useful information about $\mathbf{x}$.

\begin{lemma}\label{lemma:null_noise_suff}
A sample from
\[
p(\mathbf{z}|\mathbf{y}) = \mathcal{N}(\mathbf{A}^{+}\mathbf{y},  \beta_1 (\mathbf{I} - \mathbf{A}^{+} \mathbf{A}))
\]
is also a sufficient statistic for estimating $\mathbf{x}$ given $\mathbf{y}$, i.e.,
\[
p(\mathbf{x} | \mathbf{y}) = p(\mathbf{x} | \mathbf{z}).
\]
\end{lemma}

The null-space component of the pseudoinverse reconstruction is zero; therefore, adding null-space noise preserves sufficiency.

\begin{lemma}\label{lemma:forward_initializer}
The SDB forward process at the final time step, $p(\mathbf{x}_1|\mathbf{x}_0=\mathbf{x})$, is identically distributed to the SDB initializer $p(\mathbf{z}|\mathbf{x})$ from \cref{lemma:null_noise_suff}, assuming
\[
\lim_{t\rightarrow 1} \gamma_t = 1, \quad \lim_{t\rightarrow 1} \frac{\alpha_t^2}{\beta_t} = 0.
\]
\end{lemma}

Both are conditional Gaussian random vectors with mean $\mathbf{A}^{+} \mathbf{A} \mathbf{x}$ and covariance $\mathbf{A}^{+} \Sigma {\mathbf{A}^{+}}^\top + \beta_1 (\mathbf{I} - \mathbf{A}^{+} \mathbf{A})$.

\begin{lemma}\label{lemma:trained_sampler}
The trained SDB model is a principled sampler of $p(\mathbf{x}_0 | \mathbf{x}_1)$. 
\end{lemma}

Since SDB is a subset of score-based generative models with a standard Markovian forward process, the reverse SDE with a learned score network asymptotically samples from $p(\mathbf{x}_0|\mathbf{x}_1)$~\citep{anderson1982reverse}, approaching the exact score with sufficient data and model capacity.

The procedure to generate principled posterior samples from $p(\mathbf{x}|\mathbf{y})$ is as follows:
\begin{enumerate}
    \item Compute the pseudoinverse reconstruction $\mathbf{A}^{+}\mathbf{y}$.
    \item Sample null-space noise:
    \[
        \mathbf{z} \sim p(\mathbf{z}|\mathbf{y}) = \mathcal{N}(\mathbf{A}^{+}\mathbf{y},  \beta_1 (\mathbf{I} - \mathbf{A}^{+} \mathbf{A})).
    \]
    \item Sample the SDB reverse process:
    \[
        \mathbf{x}_0 \sim p(\mathbf{x}_0 | \mathbf{x}_1 = \mathbf{z}).
    \]
\end{enumerate}

By \cref{lemma:pinv_suff,lemma:null_noise_suff}, $\mathbf{z}$ is a sufficient statistic for $\mathbf{x}$. \Cref{lemma:forward_initializer} ensures that initializing the reverse process at $\mathbf{x}_1 = \mathbf{z}$ is valid. \Cref{lemma:trained_sampler} guarantees that the trained SDB model samples from $p(\mathbf{x}_0|\mathbf{x}_1)$. Combining these results, the procedure produces principled posterior samples from $p(\mathbf{x}|\mathbf{y})$. 

\hfill $\square$

\subsection{Magnus expansion generality}

The matrix integral formulation of the reverse-time SDB process holds under the condition that the drift matrices $\mathbf{F}_t$ commute for all $t \in [0,1]$. In the proposed SDB formulation (parameterized as in~\cref{eq:sdb_drift}), this condition is always satisfied: the only time-dependent parameter is the scalar $\alpha_t$, so the singular vectors of $\mathbf{F}_t$ remain constant, ensuring commutativity at all times.

This property extends to all scalar diffusion processes, including VP, VE, Flow Matching~\citep{lipmanflow}, Fourier Diffusion Models~\cite{10937272}, and Subspace Diffusion Models~\cite{jing2022subspace}. While diffusion methods with non-commuting $F_t$ matrices are currently not known to us, \cref{th:sde_to_cascade} remains valid in those cases through the Magnus expansion~\cite{kamm2021stochastic}.

\section{Extended related works}

\subsection{Diffusion models for inverse problems}

In recent years, deep neural networks have gained significant attention for solving various inverse problems from different perspectives \citep{9084378, chen2022simple}. Solving \cref{eq:measurement_system} can be interpreted as sampling from the posterior distribution $p(\mathbf{x} \mid \mathbf{y})$, and generative models that support conditional generation are naturally suited to this task. Diffusion models, in particular, have emerged as SOTA tools for this purpose, thanks to their flexible mathematical formulation and expressive data priors \citep{daras2024survey,preechakul2022diffusion,sobieski2024global}. The standard approach involves extending the diffusion process to sample from $p(\mathbf{x}_0 \mid \mathbf{y})$ at $t=0$ by applying Bayes' Theorem. The conditional score function can be decomposed as $\nabla_{\mathbf{x}_t}{\log{p(\mathbf{x}_t \mid \mathbf{y})}} = \score + \nabla_{\mathbf{x}_t}{\log{p(\mathbf{y} \mid \mathbf{x}_t)}}$, where the first term, $\score$, can be approximated using a pretrained score network $\scorenet$, and the second term models the relationship between $\mathbf{x}_t$ and the measurement $\mathbf{y}$ \citep{songscore, dhariwal2021beat}. This framework has spurred considerable progress in the field, with numerous successful methods emerging \citep{kawar2022denoising, chung2022improving, chung2023diffusion, song2023pseudoinverseguided, mardanivariational, chung2024decomposed,sobieski2025rethinking}. Since this approach does not require retraining the score network for each specific problem, we refer to these methods as \emph{unsupervised}.

Several works leverage the measurement system structure when applying pretrained diffusion models to inverse problems. \citet{wangzero} (Denoising Diffusion Null-Space Models, DDNM) restrict updates during generation to the null space component of $\mathbf{x}_t$, keeping the range part fixed. \citet{song2023pseudoinverseguided} (Pseudoinverse-Guided Diffusion Models, $\Pi \text{GDM}$) approximate the likelihood score via a vector-Jacobian product, where the score network’s Jacobian is computed with automatic differentiation, and the vector reflects the range-nullspace decomposition. \citet{Garber_2024_CVPR} (Denoising Diffusion Models with Iteratively Preconditioned Guidance, DDPG) propose a method for interpolating between pseudoinverse-based and least-squares-based conditioning.

\subsection{Diffusion and Schrödinger bridges.} 

Diffusion models, while effective for high-quality image synthesis, are limited by the simplicity of Gaussian priors for $p(\mathbf{x}_1)$. Recent developments in \emph{bridge models} \citep{sarkka2019applied} generalize the diffusion process by allowing $p(\mathbf{x}_1)$ to be an arbitrary distribution. This is especially important in image restoration tasks, where paired samples $(\mathbf{x}_0, \mathbf{x}_1)$—clean and distorted images—are available. Bridge models aim to generate samples from the posterior $p(\mathbf{x}_0 \mid \mathbf{x}_1)$ by initializing with a sample from $p(\mathbf{x}_1)$ rather than Gaussian noise. While conditioning the standard score network on $\mathbf{x}_1$ is a possible approach to achieve posterior sampling, it is often suboptimal \citep{batzolis2021conditional}.

Several methods have formulated the diffusion process as a stochastic bridge. \citet{heng2021simulating} propose a simulation-based algorithm using a fixed starting and ending point with an approximation of the true score. \citet{liu2022let} extend this with Doob's h-transform \citep{doob1984classical} to bridge distributions. Simulation-free algorithms utilizing the h-transform are presented by \citet{somnath2023aligned} and \citet{peluchetti2022nondenoising}. \citet{delbracio2023inversion} construct a Brownian Bridge for direct restoration from $\mathbf{x}_1$. More recently, \citet{zhoudenoising} introduce the DDBM framework, extending the VE and VP processes, while \citet{heconsistency} study DDBM within the consistency framework \citep{song2023consistency}. \citet{zheng2025diffusion} link DDBM to Denoising Diffusion Implicit Models \citep{song2021denoising}.

The Schrödinger Bridge (SB) problem \citep{schrodinger1932theorie}, which aligns distributions via constrained forward and reverse SDEs, has also been explored. \citet{de2021diffusion} apply Iterative Proportional Fitting (IPF) to solve the SB problem. \citet{liu20232} propose a tractable class of SBs, leading to a simulation-free algorithm (I2SB). \citet{chung2023direct} extend it with an additional guidance term. \citet{shi2024diffusion} build on IPF and introduce Iterative Markovian Fitting for SB solutions.

In a related approach, \citet{luo2023image} derive a scalar case of \cref{eq:linear_forward_sde} that incorporates the start and end points, termed the \emph{mean-reverting} SDE (IR-SDE). Through specific parameterization, they show that its score function is analytically tractable, connecting to the Ornstein-Uhlenbeck (OU) process \citep{gillespie1996exact}. \citet{yue2024image} extend this work with a generalized OU process and Doob’s h-transform (GOUB). Further extensions include \citet{luo2023refusion}, who apply IR-SDE in latent spaces, and \citet{welker22_interspeech} and \citet{richter2023speech}, who adapt similar processes for speech tasks. Recently, \citet{zhu2025unidb} unify GOUB and DDBM within the framework of stochastic optimal control.

\subsection{Baselines}

\subsubsection{Unsupervised}

We begin with describing the unsupervised baselines, which rely on solving the following scalar reverse equation:
\begin{equation}
    \diff \mathbf{x}_t = [\mathbf{f}(\mathbf{x}_t, t) - g^2(t)\nabla_{\mathbf{x}_t}\log{p(\mathbf{x}_t\mid\mathbf{y})}]\diff t + g(t)\diff \rwiener,
\end{equation}
where $\mathbf{f}$ and $g$ represent the drift and diffusion coefficients respectively, while $\mathbf{y}$ is the conditioning variable representing measurements in the inverse problem context. Unsupervised diffusion-based methods rely on decomposing the score function with Bayes' Theorem through  $\nabla_{\mathbf{x}_t}{\log{p(\mathbf{x}_t \mid \mathbf{y})}} = \score + \nabla_{\mathbf{x}_t}{\log{p(\mathbf{y} \mid \mathbf{x}_t)}}$ and approximating $\score$ with a pretrained $\scorenet$, while proposing different approaches for $\nabla_{\mathbf{x}_t}{\log{p(\mathbf{y} \mid \mathbf{x}_t)}}$. 

Diffusion Posterior Sampling (DPS, \citet{chung2023diffusion}) approximates $p(\mathbf{y}\mid \mathbf{x}_t)$ with $p(\mathbf{y}\mid\hat{\mathbf{x}}_0(\mathbf{x}_t))$, where $\hat{\mathbf{x}}_0(\mathbf{x}_t) = \mathbb{E}[\mathbf{x}_0\mid\mathbf{x}_t]=\mathbf{x}_t + g^2(t)\score$ is the Tweedie's formula \citep{robbins1992empirical}, giving an approximation of the denoised image at timestep $t$. Similarly, $\score$ is also approximated with $\scorenet$ in this case. 

Pseudoinverse-Guided Diffusion Model ($\Pi$GDM, \citet{song2023pseudoinverseguided}) proposes to approximate the loglikelihood score with $(\mathbf{y} - \mathbf{A}\hat{\mathbf{x}}_0(\mathbf{x}_t))^\top(r_t^2\mathbf{A}\mathbf{A}^\top + \sigma_{\mathbf{y}}^2\mathbf{I})^{-1}\mathbf{A}\frac{\partial\hat{\mathbf{x}}_0(\mathbf{x}_t)}{\partial\mathbf{x}_t}$, where $\sigma_{\mathbf{y}}^2\mathbf{I}$ is the measurement system covariance and $r_t^2$ is a time-dependent term that should depend on the data (Appendix A.3., \citet{song2023pseudoinverseguided}). Using automatic differentiation, one can compute $\frac{\partial\hat{\mathbf{x}}_0(\mathbf{x}_t)}{\partial\mathbf{x}_t}$. By combining it with the other terms, the entire approximation can be efficiently computed as a vector-Jacobian product.

In their basic formulation, Denoising Diffusion Null-space Models (DDNM, \citet{wangzero}) rely on approximating $p(\mathbf{x}_{t-1}\mid \mathbf{x}_t, \mathbf{y})$ using the following update rule:
\begin{equation}
    \mathbf{x}_{t-1}=\frac{\sqrt{\bar{\alpha}_{t-1}}\beta_t}{1-\bar{\alpha}_t}\hat{\mathbf{x}}_0(\mathbf{x}_t, \mathbf{y}) + \frac{\sqrt{\alpha_t}(1-\bar{\alpha}_{t-1})}{1-\bar{\alpha}_t}\mathbf{x}_t + \sigma_t\boldsymbol{\epsilon},
\end{equation}
where $\boldsymbol{\epsilon}\sim\boldsymbol{\mathcal{N}}(\mathbf{0}, \mathbf{I})$ and $\alpha_t, \beta_t$ follow the notation from the original work. This rule utilizes the range-nullspace decomposition via $\hat{\mathbf{x}}_0(\mathbf{x}_t, \mathbf{y})=\mathbf{A}^+\mathbf{y} + (\mathbf{I} - \mathbf{A}^+\mathbf{A})\hat{\mathbf{x}}_0(\mathbf{x}_t)$, which preserves the true range space component, while updating the null space part with Tweedie's estimate.

\subsubsection{Supervised}

We proceed with a description of baseline supervised bridge methods. All of these approaches assume a stochastic process conditioned on both the starting and ending point with the following forward and reverse equations:
\begin{subequations}
\begin{align}
    \diff \mathbf{x}_t =& \mathbf{f}(\mathbf{x}_t, \mathbf{x}_T, t, T)\diff t + g(t)\diff \wiener,\\
    \diff \mathbf{x}_t =& \mathbf{f}'(\mathbf{x}_t, \mathbf{x}_T, t, T)\diff t + g(t)\diff \rwiener,
\end{align}
\end{subequations}
where $\mathbf{f},\mathbf{f}'$ represent general drift coefficients respectively for the forward and reverse equation, $g$ is the diffusion coefficient and $\mathbf{x}_T$ represents the endpoint for the initial $\mathbf{x}_0$.

Image-to-Image Schrodinger Bridges (I2SB, \citet{liu20232}) formulate the distribution of $\mathbf{x}_t$, given starting and ending points $\mathbf{x}_0, \mathbf{x}_T$, as $p(\mathbf{x}_t\mid\mathbf{x}_0,\mathbf{x}_T) = \boldsymbol{\mathcal{N}}(\frac{\bar{\sigma}_t^2}{\bar{\sigma}_t^2 + \sigma_t^2}\mathbf{x}_0 + \frac{\sigma_t^2}{\bar{\sigma}_t^2 + \sigma_t^2}\mathbf{x}_T, \frac{\sigma_t^2\bar{\sigma}_t^2}{\bar{\sigma}_t^2 + \sigma_t^2}\mathbf{I})$, where $\sigma_t^2=\int_0^tg^2(\tau)\diff\tau, \bar{\sigma}_t^2=\int_t^1g^2(\tau)\diff\tau$ and show its equivalence to DDPM posterior sampling \citep{ho2020denoising}.

Image Restoration SDE (IR-SDE, \citet{luo2023image}) is based on formulating the forward and reverse equations as
\begin{subequations}
    \begin{align}
        \diff \mathbf{x}_t =& \theta_t(\mathbf{x}_T - \mathbf{x}_t)\diff t + \sigma(t)\diff \wiener,\\
        \diff \mathbf{x}_t =& [\theta_t(\mathbf{x}_T - \mathbf{x}_t) - \sigma^2(t)\score]\diff t + \sigma(t)\diff \rwiener,
    \end{align}
\end{subequations}
where $\theta_t=\frac{\sigma^2_t}{\lambda^2},\bar{\theta}_t=\int_0^t{\theta_\tau}d\tau$ and $\lambda$ is a predefined constant. At $t=1$, IR-SDE arrives at a Gaussian distribution (with non-zero covariance) centered at $\mathbf{x}_T$.

Generalized Ornstein-Uhlenbeck Bridges (GOUB, \citet{yue2024image}) show IR-SDE as a special case of their framework by incorporating the Doob's h-transform \citep{doob1984classical}, which pulls the process towards the desired endpoint. Formally, it is defined as $\mathbf{h}(\mathbf{x}_t, t, \mathbf{x}_T, T)=\nabla_{\mathbf{x}_t}\log{p(\mathbf{x}_T\mid\mathbf{x}_t)}$ and incorporated into the forward and reverse equations:
\begin{subequations}
    \begin{align}
        \diff \mathbf{x}_t=&\left((\theta_t + g^2(t) \frac{e^{-2 \bar{\theta}_{t:T}}}{\bar{\sigma}_{t:T}^{2}})(\mathbf{x}_T - \mathbf{x}_t)\right)\diff t + g(t)\diff \wiener,\\
        \diff \mathbf{x}_t=&\left((\theta_t + g^2(t) \frac{e^{-2 \bar{\theta}_{t:T}}}{\bar{\sigma}_{t:T}^{2}})(\mathbf{x}_T - \mathbf{x}_t) - g^2(t)\nabla_{\mathbf{x}_t}\log{p(\mathbf{x}_t\mid \mathbf{x}_T)}\right)\diff t + g(t)\diff \rwiener,
    \end{align}
\end{subequations}
where $\theta_t=\frac{g^2(t)}{2\lambda^2},\bar{\theta}_{s:t}=\int_s^t\theta_\tau d\tau,\bar{\sigma}_{s:t}^2=\frac{g^2(t)}{2\theta_t}(1 - e^{-2\bar{\theta}_{s:t}})$ for a predefined constant $\lambda$.

Denoising Diffusion Bridge Model, (DDBM, \citet{zhoudenoising}) also utilize the h-transform, but instead show how to adapt prior unconditional processes, which map images to Gaussian noise, to construct a bridge between arbitrary distributions given paired data:
\begin{subequations}
\begin{align}
    \diff \mathbf{x}_t =& [\mathbf{f}(\mathbf{x}_t, t) + g^2(t)\mathbf{h}(\mathbf{x}_t, t, \mathbf{x}_T, T)]\diff t + g(t)\diff \wiener,\\
    \diff \mathbf{x}_t =& [\mathbf{f}(\mathbf{x}_t, t) - g^2(t)\left(\nabla_{\mathbf{x}_t}\log{p(\mathbf{x}_t\mid \mathbf{x}_T)} - \mathbf{h}(\mathbf{x}_t, t, \mathbf{x}_T, T)\right)]\diff t + g(t)\diff \rwiener,
\end{align}
\end{subequations}
where $\mathbf{f}(\mathbf{x}_t, t), g(t)$ follow the original image-to-noise process.

\section{Extended methodology}

\subsection{Network parameterization}

In a scalar setting, various reparameterizations of the score function were shown to provide different trade-offs in the final performance \citep{ho2020denoising}. For example, the network could instead predict the added noise directly, which is bijectively obtained from the score. As these reparameterizations are related through simple scalar functions, one can choose them freely without additional considerations. In the matrix-valued setting, however, the choice of a parameterization is more subtle. In this case, observe that $\score=-\boldsymbol{\Sigma}_t^{-1}(\mathbf{x}_t - \mathbf{H}_t\mathbb{E[}\mathbf{x}_0 \mid \mathbf{x}_t])$. Training a score-prediction model hence requires access to the inverse of $\boldsymbol{\Sigma}_t$, which may in general be costly to obtain. By properly choosing the form of $\gamma_t$ and $\beta_t$, one can show that $\mathbf{G}_t\mathbf{G}_t^\top\score=(f_t\mathbf{I} - 2\mathbf{F}_t)(\mathbf{H}_t\mathbb{E}[\mathbf{x}_0 \mid \mathbf{x}_t] - \mathbf{x}_t)$ for some function $f_t$, \ie, an $\mathbf{x}_0$-prediction model alleviates the need for computing $\boldsymbol{\Sigma}_t^{-1}$. Therefore, we treat this parameterization as the default one for SDB.

\subsection{Nonlinear measurement systems}\label{sec:method_nonlinear}

While SDB is primarily formulated for linear operators $\mathbf{A}$ with a global range-nullspace decomposition, it can be extended to certain nonlinear inverse problems. For a nonlinear, differentiable system response $\mathcal{\mathbf{A}}$, we can approximate it locally using a first-order Taylor expansion around some point $\mathbf{x}_0$:
\begin{equation}
\label{eq:nonlinear_linearization}
\mathcal{\mathbf{A}}(\mathbf{x}) \approx \mathcal{\mathbf{A}}(\mathbf{x}_0) + \mathcal{\boldsymbol{J}}_{\mathcal{\mathbf{A}}}(\mathbf{x}_0) (\mathbf{x} - \mathbf{x}_0),
\end{equation}

where $\mathcal{\boldsymbol{J}}_{\mathcal{\mathbf{A}}}(\mathbf{x}_0)$ is the Jacobian of $\mathcal{\mathbf{A}}$ at $\mathbf{x}_0$. This linearization induces a local range-nullspace decomposition, enabling the use of SDB on nonlinear systems. The procedure is as follows:
\begin{enumerate}
    \item For a measurement $\mathbf{y} = \mathcal{\mathbf{A}}(\mathbf{x}) + \boldsymbol{\Sigma}^{1/2}\epsilon$, compute an approximate maximum likelihood estimate $\hat{\mathbf{x}}$ of the signal $\mathbf{x}$.
    \item Linearize the operator at $\hat{\mathbf{x}}$ using \cref{eq:nonlinear_linearization} and compute the Jacobian $\mathbf{J}_{\mathbf{A}}(\hat{\mathbf{x}})$.
    \item Treat $\mathbf{y} \approx \mathbf{J}_{\mathbf{A}}(\hat{\mathbf{x}})\mathbf{x} + \boldsymbol{\Sigma}^{1/2}\epsilon$ as a linear Gaussian model and apply standard SDB with pseudoinverse and range/nullspace projections.
\end{enumerate}

This approach leverages the Jacobian as a local linear approximation of the nonlinear operator, effectively adapting SDB to more general, differentiable systems.

\section{Experimental setup}

\subsection{Training hyperparameters}

We follow the training procedure proposed by \citet{luo2023image}, using the ADAM optimizer \citep{kingma2017adammethodstochasticoptimization} with an initial learning rate of $1 \times 10^{-4}$, no weight decay, and $(\beta_1, \beta_2) = (0.9, 0.99)$. A multi-step learning rate scheduler is applied, halving the learning rate at the 36th, 60th, 72nd, and 90th epochs, as in the original work. All methods are trained using the $\ell_1$ loss function with a batch size of 8. To ensure fairness across methods, we evaluate each model every 16 epochs and report the performance of the best checkpoint, rather than relying solely on the final one.

\subsection{Computational requirements}

All experiments were conducted on a cluster of NVIDIA A100 GPUs, with each method trained using a single GPU. The approximate training times for each task are as follows: 4 hours for super-resolution, 24 hours for MRI reconstruction, 48 hours for CT reconstruction, and 72 hours for inpainting.

\section{Additional experimental results}

\subsection{Evaluation under a misspecified model (CT reconstruction)}

\begin{figure}[h]
    \centering
    \includegraphics[width=1.0\linewidth]{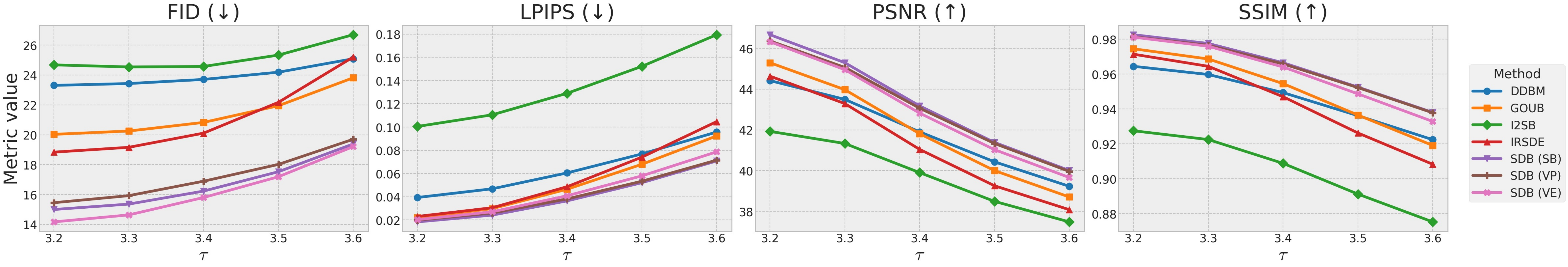}
    \includegraphics[width=1.0\linewidth]{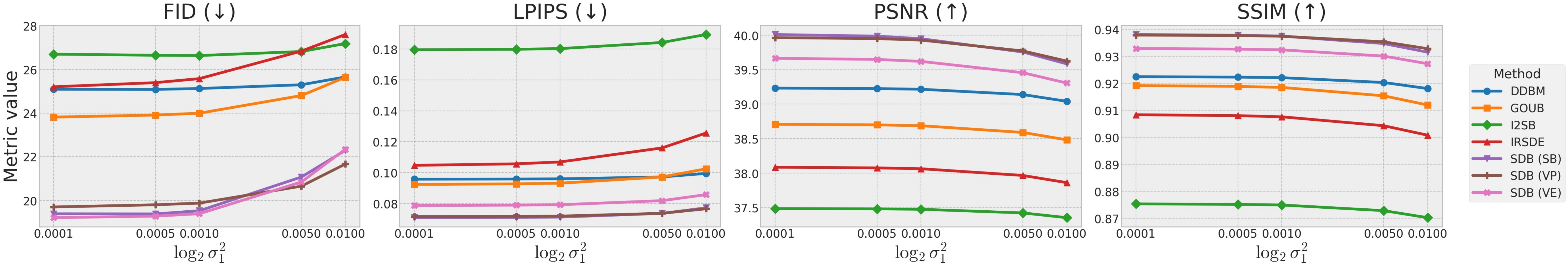}
    \caption{Quantitative comparison of SDB (SB) with other bridge methods in a misspecified CT reconstruction setting. The \textbf{top} row evaluates bridges trained with $\tau = 3.2$, $\sigma_1^2 = 0.0001$ on data generated from systems with increasing $\tau$. The \textbf{bottom} row evaluates performance on data with $\tau=3.6$ and increasing $\sigma_1^2$. Perturbing the original system makes the problem harder in both cases.}
    \label{fig:ct_misspecificed_model}
\end{figure}

\Cref{fig:ct_misspecificed_model} presents the results of an analogous analysis of performance of bridge methods under a misspecified model (here in CT reconstruction task), where the first part of the experiment considers the default setting of $\sigma_1^2$ with increasing $\tau$, while the second part shows the results for $\tau=3.6$ and increasing $\sigma_1^2$. In a similar manner to \cref{sec:misspecified_model}, all SDB variants achieve a clear advantage over other bridge methods when the system's parameters are perturbed.

\subsection{More challenging misspecified settings}

We investigate more challenging and realistic misspecified scenarios for supervised inverse problems, including adversarial modifications of the forward operator and noise model. We evaluate SDB (SB) against supervised baselines across these settings.

\subsubsection{Misspecified operator}

We consider two superresolution scenarios: a drastically enlarged downsampling kernel ($32\times32$) and JPEG compression at 30\% quality. Both create inputs very different from training-time PRs. \Cref{tab:misspecified_operator} reports results across both tasks, with the task indicated in the first column. SDB (SB) consistently achieves the best performance in 3 out of 4 metrics.

\begin{table}[h]
\centering
\caption{Superresolution under more challenging operators (bold indicates best values).}
\label{tab:misspecified_operator}
\begin{tabular}{lcccc|cccc}
\toprule
Method & \multicolumn{4}{c|}{Kernel $32\times32$} & \multicolumn{4}{c}{JPEG $30\%$} \\
\cmidrule(r){2-5} \cmidrule(r){6-9}
       & FID↓ & LPIPS↓ & PSNR↑ & SSIM↑ & FID↓ & LPIPS↓ & PSNR↑ & SSIM↑ \\
\midrule
IRSDE      & 130.47 & 0.684 & 17.85 & 0.381 & 99.43  & \textbf{0.283} & 24.81 & 0.651 \\
I2SB       & 131.33 & \textbf{0.681} & 17.84 & 0.381 & 93.62  & 0.322 & 25.12 & 0.668 \\
DDBM       & 131.55 & 0.700 & 17.83 & 0.380 & 92.39  & 0.373 & 25.19 & 0.685 \\
GOUB       & 134.66 & 0.690 & 17.85 & 0.370 & 101.08 & 0.291 & 24.91 & 0.662 \\
SDB (SB)   & \textbf{129.87} & 0.682 & \textbf{17.86} & \textbf{0.390} & \textbf{90.26} & 0.380 & \textbf{25.41} & \textbf{0.692} \\
\bottomrule
\end{tabular}
\end{table}

\subsubsection{Misspecified noise model}

We also consider a misspecified noise model, replacing the Gaussian noise with a Poisson distribution to simulate photon-counting physics. For a given signal sample $\mathbf{x}$, measurements are sampled as $\mathbf{y} \sim \text{Poisson}(\mathbf{I}_0 \cdot \exp(-\mathbf{A}\mathbf{x}))$, where $\mathbf{I}_0$ is the incident photon intensity. We evaluate CT reconstruction, where Poisson noise closely reflects real-world X-ray physics. As shown in \cref{tab:misspecified_noise_ct}, SDB outperforms all supervised baselines across different $\mathbf{I}_0$ values, even without explicitly modeling the Poisson distribution.

\begin{table}[h]
\centering
\caption{CT reconstruction with Poisson noise. Bold indicates best values.}
\label{tab:misspecified_noise_ct}
\tiny
\begin{tabular}{lcccc|cccc|cccc}
\toprule
Method & \multicolumn{4}{c|}{$\mathbf{I}_0=10000$} & \multicolumn{4}{c|}{$\mathbf{I}_0=5000$} & \multicolumn{4}{c}{$\mathbf{I}_0=1000$} \\
\cmidrule(r){2-5} \cmidrule(r){6-9} \cmidrule(r){10-13}
& FID↓ & LPIPS↓ & PSNR↑ & SSIM↑ & FID↓ & LPIPS↓ & PSNR↑ & SSIM↑ & FID↓ & LPIPS↓ & PSNR↑ & SSIM↑ \\
\midrule
GOUB      & 17.78 & 0.0230 & 44.85 & 0.9717 & 17.83 & 0.0240 & 44.59 & 0.9705 & 20.52 & 0.0491 & 42.79 & 0.9588 \\
IRSDE     & 17.29 & 0.0256 & 44.17 & 0.9681 & 17.86 & 0.0299 & 43.88 & 0.9665 & 21.12 & 0.0678 & 41.84 & 0.9506 \\
DDBM      & 20.69 & 0.0402 & 44.11 & 0.9622 & 20.57 & 0.0406 & 43.93 & 0.9613 & 21.75 & 0.0575 & 42.53 & 0.9525 \\
I2SB      & 21.91 & 0.1014 & 41.68 & 0.9242 & 21.82 & 0.1018 & 41.54 & 0.9231 & 22.43 & 0.1170 & 40.44 & 0.9130 \\
SDB (SB)  & \textbf{14.03} & \textbf{0.0193} & \textbf{46.05} & \textbf{0.9799} & \textbf{14.61} & \textbf{0.0203} & \textbf{45.70} & \textbf{0.9785} & \textbf{19.80} & \textbf{0.0461} & \textbf{43.44} & \textbf{0.9657} \\
\bottomrule
\end{tabular}
\end{table}

We further consider MRI reconstruction under Poisson noise as an adversarial setting, where no natural process is typically modeled with Poisson noise, making it highly out-of-distribution. \Cref{tab:misspecified_noise_mri} shows that SDB again outperforms other methods in 3 out of 4 metrics while almost matching the best SSIM.

\begin{table}[h]
\centering
\caption{MRI reconstruction with Poisson noise (bold indicates best values).}
\label{tab:misspecified_noise_mri}
\small
\begin{tabular}{lcccc|cccc}
\toprule
Method & \multicolumn{4}{c|}{$\mathbf{I}_0=20.0$} & \multicolumn{4}{c}{$\mathbf{I}_0=16.0$} \\
\cmidrule(r){2-5} \cmidrule(r){6-9}
& FID↓ & LPIPS↓ & PSNR↑ & SSIM↑ & FID↓ & LPIPS↓ & PSNR↑ & SSIM↑ \\
\midrule
DDBM      & 40.2161 & 0.2647 & 21.9509 & \textbf{0.5572} & 39.0941 & 0.2360 & 22.6901 & \textbf{0.5919} \\
GOUB      & 39.8257 & 0.2821 & 21.4963 & 0.5246 & 39.3272 & 0.2530 & 22.2050 & 0.5577 \\
IRSDE     & 39.8516 & 0.2706 & 21.5772 & 0.5383 & 39.4095 & 0.2445 & 22.2666 & 0.5722 \\
I2SB      & 40.0352 & 0.2767 & 21.4514 & 0.5219 & 39.0571 & 0.2492 & 22.1715 & 0.5546 \\
SDB (SB)  & \textbf{39.7303} & \textbf{0.2513} & \textbf{22.0684} & 0.5569 & \textbf{38.9677} & \textbf{0.2245} & \textbf{22.8237} & 0.5891 \\
\midrule
Method & \multicolumn{4}{c|}{$\mathbf{I}_0=12.0$} & \multicolumn{4}{c}{$\mathbf{I}_0=8.0$} \\
\cmidrule(r){2-5} \cmidrule(r){6-9}
& FID↓ & LPIPS↓ & PSNR↑ & SSIM↑ & FID↓ & LPIPS↓ & PSNR↑ & SSIM↑ \\
\midrule
DDBM      & 38.2960 & 0.2036 & 23.6083 & \textbf{0.6345} & 36.6739 & 0.1653 & 24.7777 & \textbf{0.6873} \\
GOUB      & 38.4142 & 0.2196 & 23.0794 & 0.5991 & 36.9538 & 0.1794 & 24.2220 & 0.6524 \\
IRSDE     & 38.4208 & 0.2141 & 23.1378 & 0.6139 & 37.0698 & 0.1769 & 24.2971 & 0.6680 \\
I2SB      & \textbf{38.1304} & 0.2168 & 23.0778 & 0.5957 & 37.1202 & 0.1782 & 24.2578 & 0.6492 \\
SDB (SB)  & 38.1484 & \textbf{0.1936} & \textbf{23.7615} & 0.6293 & \textbf{36.6346} & \textbf{0.1569} & \textbf{24.9707} & 0.6811 \\
\bottomrule
\end{tabular}
\end{table}

\subsection{Motion deblurring with an ill-conditioned system}

To assess the practicality of SDB in ill-conditioned systems, we consider a motion deblurring task on $128 \times 128$ flower images from the Flowers102 dataset~\citep{nilsback2008automated}. The system matrix is modeled as a block Toeplitz matrix with Toeplitz blocks (BTTB), implemented via 2D convolutions in the pixel domain. For the pseudoinverse approximation, we apply zero-order Tikhonov regularization (Wiener filtering) in the frequency domain. Although this differs from the exact pseudoinverse, SDB matches the best performance in LPIPS and clearly surpasses all supervised baselines in the other three metrics, demonstrating robustness to ill-conditioned operators (\cref{tab:motion_deblur}).

\begin{table}[h]
\centering
\caption{Motion deblurring performance on Flowers102 dataset using BTTB system matrix with Tikhonov-regularized pseudoinverse. Bold indicates best values.}
\label{tab:motion_deblur}
\begin{tabular}{lcccc}
\toprule
Method & FID↓ & LPIPS↓ & PSNR↑ & SSIM↑ \\
\midrule
GOUB      & 17.03 & 0.053 & 23.84 & 0.813 \\
IRSDE     & 16.27 & 0.033 & 26.76 & 0.842 \\
DDBM      & 16.00 & 0.040 & 29.74 & 0.892 \\
I2SB      & 15.62 & \textbf{0.025} & 29.29 & 0.874 \\
SDB (SB)  & \textbf{14.89} & \textbf{0.025} & \textbf{30.39} & \textbf{0.903} \\
\bottomrule
\end{tabular}
\end{table}

\subsection{Contrast recovery with a nonlinear system}

In the following, we provide an initial study concerning the application of SDB to nonlinear systems.

In practice, the initial estimate $\hat{\mathbf{x}}$ of the true signal $\mathbf{x}$ mentioned in \cref{sec:method_nonlinear} can be obtained via a few iterations of gradient descent on
\[
\min_{\hat{\mathbf{x}}} \|\mathbf{A}(\hat{\mathbf{x}}) - \mathbf{y}\|_2^2,
\]
providing a reasonable starting point for linearization. Steps 2 and 3 (\cref{sec:method_nonlinear}) involve computing the Jacobian at the given point and using it to obtain the pseudoinverse reconstruction (PR) and range-nullspace projections for SDB. These operations can be implemented efficiently without explicitly storing the Jacobian, using Jacobian-vector products (JVPs) or gradient-based operations available in autodifferentiation frameworks such as PyTorch, or by exploiting an analytic expression for the Jacobian when available.

To illustrate the applicability of SDB to nonlinear systems, we perform a proof-of-concept benchmark on CIFAR10~\citep{krizhevsky2009learning} using a nonlinear contrast operator
\[
\mathbf{A}(\mathbf{x}) = \sigma(k (\mathbf{x} - \boldsymbol{\alpha})),
\]
where $\sigma$ is the sigmoid function, $k$ controls contrast strength, and $\boldsymbol{\alpha}$ is a contrast bias. We apply the linearization procedure described in \cref{sec:method_nonlinear}: step 1 uses 5 iterations of gradient descent to obtain an initial guess, while steps 2 and 3 use an analytic Jacobian to compute the pseudoinverse reconstruction and projections.

As shown in \cref{tab:nonlinear_cifar}, SDB (SB) outperforms all supervised baselines across all metrics in this nonlinear contrast recovery task. This demonstrates that, even under nonlinear system responses, SDB can effectively leverage the locally linearized operator to produce high-fidelity reconstructions, achieving both superior perceptual quality (FID, LPIPS) and reconstruction accuracy (PSNR, SSIM). We emphasize that this represents only an initial proof-of-concept study, and future work should investigate the applicability of SDB to nonlinear systems more thoroughly.

\begin{table}[h]
\centering
\caption{Nonlinear contrast reconstruction on CIFAR10. Bold indicates best values.}
\label{tab:nonlinear_cifar}
\begin{tabular}{lcccc}
\toprule
Method & FID↓ & LPIPS↓ & PSNR↑ & SSIM↑ \\
\midrule
I2SB      & 2.97 & 0.00103 & 37.53 & 0.9906 \\
GOUB      & 4.60 & 0.00182 & 32.54 & 0.9743 \\
DDBM      & 1.92 & 0.00016 & 42.74 & 0.9963 \\
IRSDE     & 4.35 & 0.00140 & 33.53 & 0.9789 \\
SDB (SB)  & \textbf{1.31} & \textbf{0.00015} & \textbf{44.48} & \textbf{0.9977} \\
\bottomrule
\end{tabular}
\end{table}

\subsection{Hyperparameter analysis}

\begin{figure}
    \centering
    \includegraphics[width=1.0\linewidth]{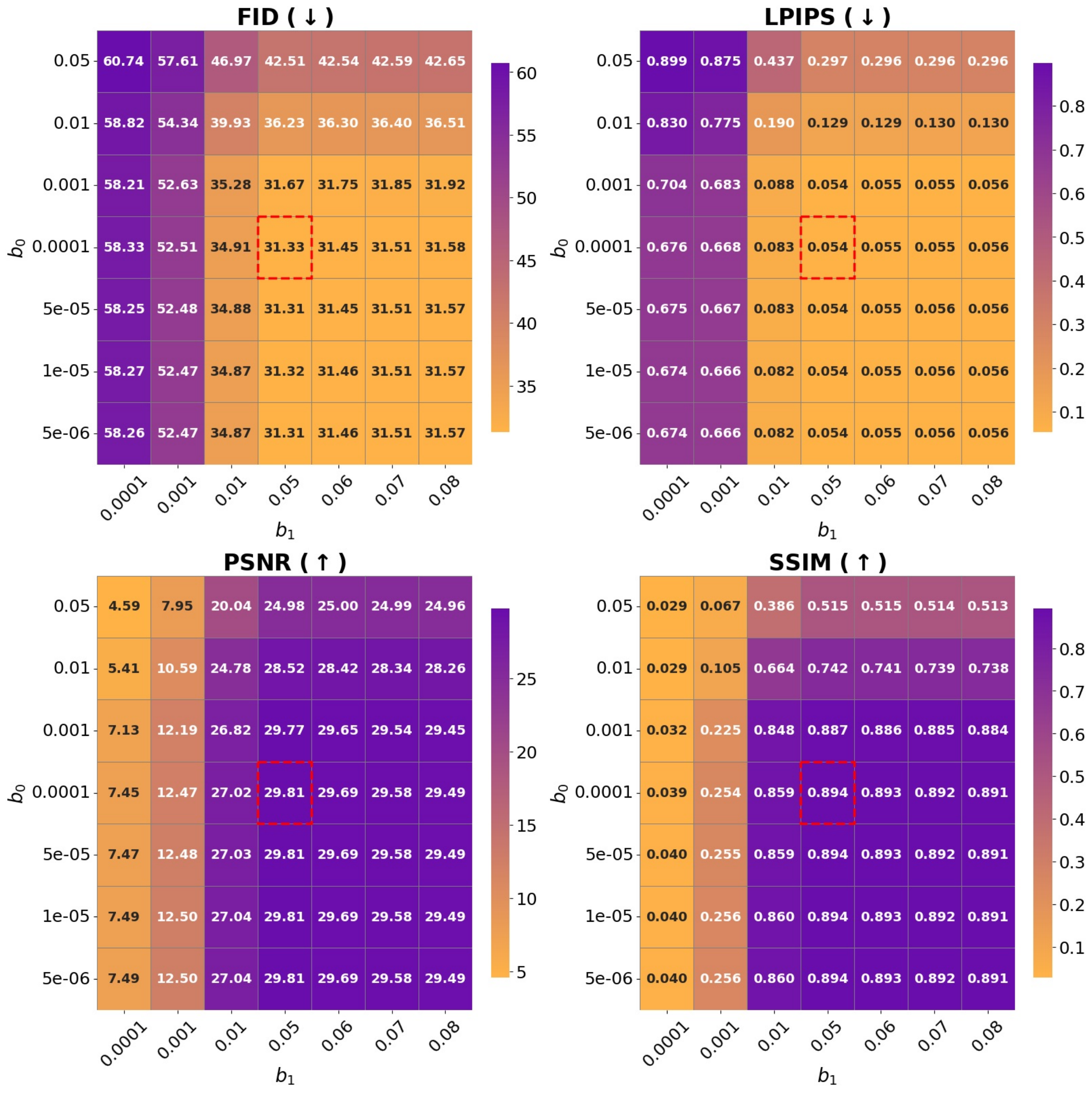}
    \includegraphics[width=1.0\linewidth]{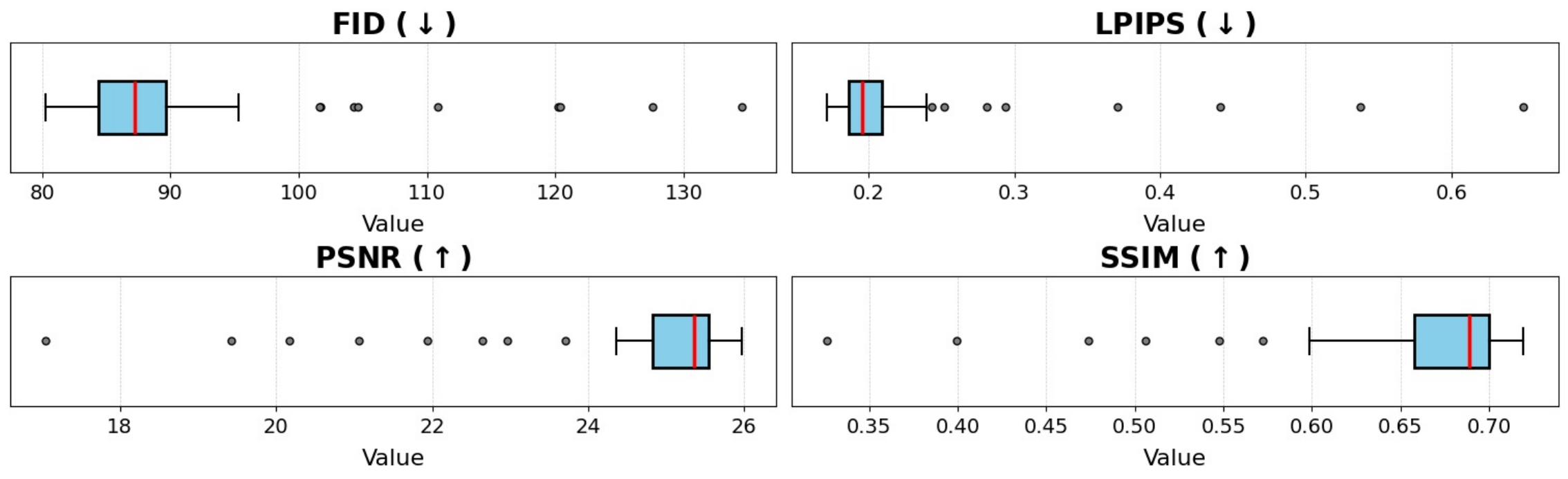}
    \caption{\textbf{Top}: Performance heatmaps for MRI reconstruction under varying $(\epsilon_1, \epsilon_2)$, using the optimal $(b_0, b_1)$ configuration. The red box highlights the LPIPS-optimal setting. \textbf{Bottom}: Boxplots showing the best achievable performance across 64 $(b_0, b_1)$ pairs on the superresolution task, where $(\epsilon_1, \epsilon_2)$ are tuned separately for each $(b_0, b_1)$ pair.}
    \label{fig:hp_analysis}
\end{figure}

We conduct an ablation study to analyze the sensitivity of SDB to its hyperparameters. Following prior works \citep{zhoudenoising}, our stochastic process is undefined at $t = 1$, requiring sampling to begin at $t = 1 - \epsilon_1$ for some $\epsilon_1 > 0$. Similarly, we terminate sampling at $t = \epsilon_2$, where $\epsilon_2 > 0$. For SDB (VP) and SDB (VE), $\epsilon_1$ and $\epsilon_2$ are the only hyperparameters. In contrast, SDB (SB) includes an additional design choice: the shape of the null-space diffusion coefficient $g(t)$. We parameterize it as
\[
g^2(t) = \mathbbm{1}_{t \leq 0.5} \hat{g}(t) + \mathbbm{1}_{t > 0.5} \hat{g}(1 - t)
\]
where $\hat{g}(t) = (\sqrt{b_0} + t(\sqrt{b_1} - \sqrt{b_0}))^2$, which is a continuous version of the coefficient proposed by \citet{liu20232}. This introduces two additional hyperparameters, $b_0, b_1 > 0$. In the remainder of this section, we focus on SDB (SB), noting that we observed qualitatively similar trends for the VP and VE variants.

We first examine how the choice of $\epsilon_1$ and $\epsilon_2$ influences the optimal configuration of $(b_0, b_1)$ for SDB (SB) in the MRI reconstruction task. \Cref{fig:hp_analysis} (top) shows the performance trend when varying $(\epsilon_1, \epsilon_2)$. We observe a simple and stable relationship with performance, where $\epsilon_1$ has a stronger influence than $\epsilon_2$. Moreover, a broad plateau emerges, allowing for easy tuning. We also highlight the best configuration in terms of LPIPS, which lies close to the optimal Pareto frontier in this setting.

Next, we analyze the stability of SDB (SB) across different $(b_0, b_1)$ configurations. Since these values determine the shape of the variance function, the optimal $\epsilon_1, \epsilon_2$ may differ across $(b_0, b_1)$ pairs. \Cref{fig:hp_analysis} (bottom) presents boxplots of the best performance for 64 configurations of $(b_0, b_1)$ on the superresolution task. For each pair, we use grid search to identify the best $\epsilon_1, \epsilon_2$ values, leveraging the small size of the DIV2K dataset. The results reveal strong stability, with performance concentrated around the best observed value and only a few outliers.

\subsection{Ablation study on noise schedules}

We evaluate the effect of different schedules for the VP and VE variants of SDB, considering linear, quadratic, and cosine versions on the superresolution task. Using our notation: 
\[
\alpha_t = 1 - t^2 \quad \text{(quadratic VP)}, \quad 
\alpha_t = \cos\left(\frac{\pi}{2} t\right) \quad \text{(cosine VP)}, 
\] 
\[
\beta_t = \sigma_{\max} t^2 \quad \text{(quadratic VE)}, \quad 
\beta_t = \sigma_{\max} \sin\left(\frac{\pi}{2} t\right) \quad \text{(cosine VE)}.
\]
We also assume $\gamma_t = \sigma_{\max}^{-1} \beta_t$ for VE.

\begin{table}[h]
\centering
\caption{Ablation results for SDB (VP and VE) with different noise schedules. Lower FID and LPIPS indicate better perceptual quality; higher PSNR and SSIM indicate better reconstruction fidelity. Bold indicates the best value for each metric within VP or VE.}
\label{tab:ablation_schedules}
\setlength{\tabcolsep}{6pt}
\renewcommand{\arraystretch}{1.1}
\begin{tabular}{l|cccc|cccc}
\toprule
 & \multicolumn{4}{c|}{SDB (VP)} & \multicolumn{4}{c}{SDB (VE)} \\
\cmidrule(lr){2-5} \cmidrule(lr){6-9}
Schedule & FID ↓ & LPIPS ↓ & PSNR ↑ & SSIM ↑ & FID ↓ & LPIPS ↓ & PSNR ↑ & SSIM ↑ \\
\midrule
Linear    & 87.08 & 0.228 & \textbf{25.91} & \textbf{0.724} & 94.73 & 0.226 & 25.90 & 0.718 \\
Quadratic & \textbf{75.25} & 0.174 & 24.77 & 0.680 & 76.88 & \textbf{0.159} & 25.29 & 0.678 \\
Cosine    & 77.99 & \textbf{0.160} & 24.75 & 0.674 & \textbf{76.14} & 0.217 & \textbf{26.17} & \textbf{0.732} \\
\bottomrule
\end{tabular}
\end{table}

The results show a clear trade-off between perceptual quality (FID, LPIPS) and reconstruction fidelity (PSNR, SSIM). Linear schedules favor distortion metrics, while quadratic and cosine schedules improve perceptual metrics at the expense of reconstruction accuracy. In particular, Cosine SDB (VE) achieves the best values on three out of four metrics, demonstrating that exploring alternative schedules can further improve performance.

\subsection{Computational complexity and overhead of SDB}

We analyze the computational complexity and runtime overhead of SDB compared to other supervised baselines. First, we discuss the asymptotic complexity, followed by empirical measurements.

\subsubsection*{Complexity}

For the general case where the pseudoinverse can be computed analytically using a stored matrix $\mathbf{A}$, two cases arise: $m \geq d$ and $m < d$. The pseudoinverse can be computed as
\[
\mathbf{A}^+ = (\mathbf{A}^\top \mathbf{A})^{-1} \mathbf{A}^\top \quad \text{or} \quad \mathbf{A}^+ = \mathbf{A}^\top (\mathbf{A} \mathbf{A}^\top)^{-1},
\] 
respectively. The resulting asymptotic complexities are summarized in \cref{tab:complexity}. Computing the PR has the same complexity (in the $O$ sense) as computing the pseudoinverse itself.

\begin{table}[h]
\centering
\caption{Asymptotic computational complexity for pseudoinverse computation and range/null space projections.}
\label{tab:complexity}
\begin{tabular}{lcc}
\toprule
Operation Type & $m \geq d$ & $m < d$ \\
\midrule
Pseudoinverse & $O(md^{2} + d^{3})$ & $O(m^{2}d + m^{3})$ \\
Range Projection $\mathbf{A}\mathbf{A}^{+}$ & $O(md^{2})$ & $O(m^{2}d)$ \\
Null Projection $(\mathbf{I} - \mathbf{A}^{+}\mathbf{A})$ & $O(d^{2}m)$ & $O(dm^{2})$ \\
\bottomrule
\end{tabular}
\end{table}

In practice, we do not explicitly construct $\mathbf{A}^+$. Instead, we implement the application of $\mathbf{A}$, $\mathbf{A}^\top$, $\mathbf{A}^+$, and their combinations to obtain projections and PRs. For each inverse problem, these implementations differ. For example, in inpainting, $\mathbf{A}$ corresponds to element-wise masking, whereas in CT reconstruction, truncated SVD is used to implement $\mathbf{A}$ efficiently, typically resulting in lower empirical complexity.

\subsubsection*{Empirical Overhead}

We measured runtime in seconds for the most computationally intensive operations on the CT reconstruction task. The evaluation was conducted over four batches of size eight. We report three types of operations: 

\begin{itemize}
\item Forward Step Without Network: sampling from $p(\mathbf{x}_t \mid \mathbf{x}_0)$,
\item Forward Step With Network: Forward Step Without Network plus a forward pass through the neural network,
\item Reverse Step With Network: sampling from $p(\mathbf{x}_{t-\Delta t} \mid \mathbf{x}_t, \mathbf{x}_1)$ using the neural network, corresponding to a single discrete reverse-time step.
\end{itemize}

The empirical runtimes are summarized in \cref{tab:runtime}.

\begin{table}[h]
\centering
\caption{Runtime of SDB and baseline supervised methods for the CT reconstruction task (seconds per batch of size 8).}
\label{tab:runtime}
\begin{tabular}{lccc}
\toprule
Method & \scalebox{0.85}{Forward Step Without Network} & \scalebox{0.85}{Forward Step With Network} & \scalebox{0.85}{Reverse Step With Network} \\
\midrule
SDB (SB) & 0.0062 & 0.0885 & 0.5981 \\
I2SB     & 0.0003 & 0.0821 & 0.5851 \\
IR-SDE   & 0.0003 & 0.0832 & 0.5895 \\
GOUB     & 0.0008 & 0.0830 & 0.5863 \\
DDBM     & 0.0011 & 0.0825 & 0.5880 \\
\bottomrule
\end{tabular}
\end{table}

While Forward Step With Network shows a small increase in computation time for SDB compared with other methods, this overhead is negligible in practice. The relative cost of the neural network forward pass dominates runtime, and matrix-based operations are efficiently parallelized on the NVIDIA A100 GPU used for all experiments. Consequently, SDB runtime during inference is nearly identical to that of other supervised methods.

\subsection{OT unpaired-data-based baselines}

We provide an additional small-scale comparison with recent OT-based unpaired-data methods (UNSB by \citet{kim2024unpaired} and SBF by \citet{de2024schrodinger}) on the MRI reconstruction task, as shown in \cref{tab:mri_extended}. Following the original training procedures, we trained the models to map between the distributions of PRs and true signal observations. Since these methods lack direct supervision from paired data, their performance is naturally inferior to that of supervised approaches.

\begin{table}[h]
\centering
\caption{MRI reconstruction results across all supervised and unpaired-data-based methods. Lower FID and LPIPS indicate better perceptual quality; higher PSNR and SSIM indicate better reconstruction fidelity.}
\label{tab:mri_extended}
\setlength{\tabcolsep}{5pt}
\renewcommand{\arraystretch}{1.1}
\resizebox{0.95\textwidth}{!}{%
\begin{tabular}{l|cc|ccccccc}
\toprule
& \multicolumn{2}{c|}{Unpaired-data-based} & \multicolumn{7}{c}{Supervised} \\
\cmidrule(lr){2-3} \cmidrule(lr){4-10}
Metric & UNSB & SBF & I2SB & IR-SDE & GOUB & DDBM & SDB (VP) & SDB (VE) & SDB (SB) \\
\midrule
FID ↓   & 35.91 & 43.63 & 31.54 & \underline{30.14} & 30.63 & 32.42 & 32.88 & 33.90 & \textbf{29.85} \\
LPIPS ↓ & 0.237 & 0.290 & 0.065 & 0.065 & \underline{0.058} & 0.074 & 0.068 & 0.083 & \textbf{0.053} \\
PSNR ↑  & 18.72 & 18.46 & 28.75 & 28.88 & 28.59 & 28.97 & \underline{29.26} & 29.10 & \textbf{29.81} \\
SSIM ↑  & 0.483 & 0.291 & 0.849 & 0.871 & 0.863 & 0.872 & \underline{0.881} & 0.876 & \textbf{0.893} \\
\bottomrule
\end{tabular}
}
\end{table}

\section{Extended discussion}

\subsection{Advantages of Embedding Measurement System Information}

Incorporating measurement system information into the generative process offers several conceptual and practical advantages. While earlier approaches that relied on hand-crafted priors often provided limited gains, the proposed SDB framework does not depend on such priors. Instead, SDB embeds the measurement system directly into the model dynamics, which naturally decomposes the inverse problem into two orthogonal components with distinct roles: the range space and the null space. This decomposition provides a structured way to separate reconstruction from synthesis without manually injecting prior knowledge.

By embedding the measurement operator, SDB defines a more efficient stochastic path from the observed measurements to the clean image, akin to other diffusion bridge models that refine conditional diffusion by employing more informed reverse-process initializations. Within this framework, the neural network simultaneously learns two complementary tasks: (i) an optional denoising task in the range space—typically simpler and often directly constrained by the data—and (ii) a synthesis task in the null space, which accounts for missing or unobserved components.

In noise-free scenarios, this structure allows exact preservation of range-space content, ensuring that information already supported by the measurements remains unaltered. Under noisy conditions, the model benefits from the fact that the range space already encodes the correct underlying structure, requiring only localized refinement. Consequently, the range space acts as a structurally informative prior that guides the null-space synthesis. For example, in generative inpainting tasks—where the unmasked region is noise-free or nearly so—SDB avoids unnecessary corruption of the observed region and concentrates its modeling capacity on generating the missing content.

\subsection{Perception-Distortion Tradeoff}

The trade-off between perception and distortion mentioned by \citet{blau2018perception} is an important topic in the context of this work. Thanks to a specific formalization, where perception is measured as a divergence between the true distribution and the one induced by the model, while distortion is quantified through the expected reconstruction error, one may hypothesize about the behavior of a given algorithm from a theoretical point of view. Following the formulation in \citet{blau2018perception}, since perception and distortion are related through a Pareto front, one can draw more principled conclusions from quantitative results.

The range-nullspace decomposition provided by the assumed system response matrix also enables deeper insights. In comparison to other methods, SDB's advantage lies in its differential treatment of the range and null spaces: it must only (optionally) denoise the former, while new content is synthesized in the latter. This dual treatment leads to improved performance from the outset. In the noise-free case, the range space component is left untouched, resulting in zero distortion, while in the noisy case it undergoes milder transformations to reach the final result, which intuitively should also reduce distortion. Under a noise-free scenario, it is also easier to reason about perception: if the model synthesizes null-space content that perfectly aligns with the ground truth image, it achieves zero distortion at an inevitable cost in perception. Conversely, if it synthesizes null-space content that is fully in-distribution but deviates from the true signal, perception is maximized at the cost of distortion.

From a theoretical perspective, in the limit, SDB samples from the true conditional distribution $p(\mathbf{x}\mid\mathbf{y})$, which suggests that the method is inclined toward perfect perception. However, since we implicitly assume a non-empty null space, SDB will always lack some information required to perfectly reconstruct the true $\mathbf{x}$, mirroring the trade-off considered by \citet{blau2018perception}.

In practice, we observe a clear trend of SDB outperforming all other methods in three out of four metrics in the super-resolution task (\cref{tab:main_results}), while obtaining slightly worse results in terms of LPIPS. Following \citet{blau2018perception}, this observation leads to an interesting insight. Better FID, representing a divergence between distributions, indicates that SDB achieves better perceptual results. However, it is also superior in terms of PSNR and SSIM, which measure reconstruction accuracy (distortion). Based on \citet{blau2018perception}, one could argue that, in this task, the methods operate near the optimal performance. Since LPIPS does not quantify distributional divergence but instead measures perceptual distortion in the space of neural network representations, it illustrates how a specific type of semantic reconstruction must be sacrificed to move closer to the Pareto front. Moreover, the ablation study on noise schedulers for the VP and VE variants of SDB shows how better LPIPS values can be recovered, providing another manifestation of this trade-off.

\section{Broader impact}

By incorporating measurement system parameters into its SDE, SDB offers positive societal impacts by enabling more accurate and efficient reconstructions in critical applications such as medical imaging, remote sensing, and scientific inverse problems, thereby supporting improved diagnostics, sustainability, and broader accessibility. However, it also poses potential negative societal risks, including misuse for surveillance or de-anonymization, amplification of bias from flawed measurement models, and overreliance on plausible but incorrect outputs in high-stakes domains. Additionally, the method could be repurposed for generating deceptive content from limited data. However, SDB and prior bridge methods have not been demonstrated at industry-scale deployment levels, and in practice, they are typically suited for problem-specific applications rather than large-scale settings with vast amounts of data. This limits their potential for misuse in broader societal contexts.

\section{Additional visual results}

\begin{figure}[h]
    \hspace{-2.5em}
    \includegraphics[width=1.12\linewidth]{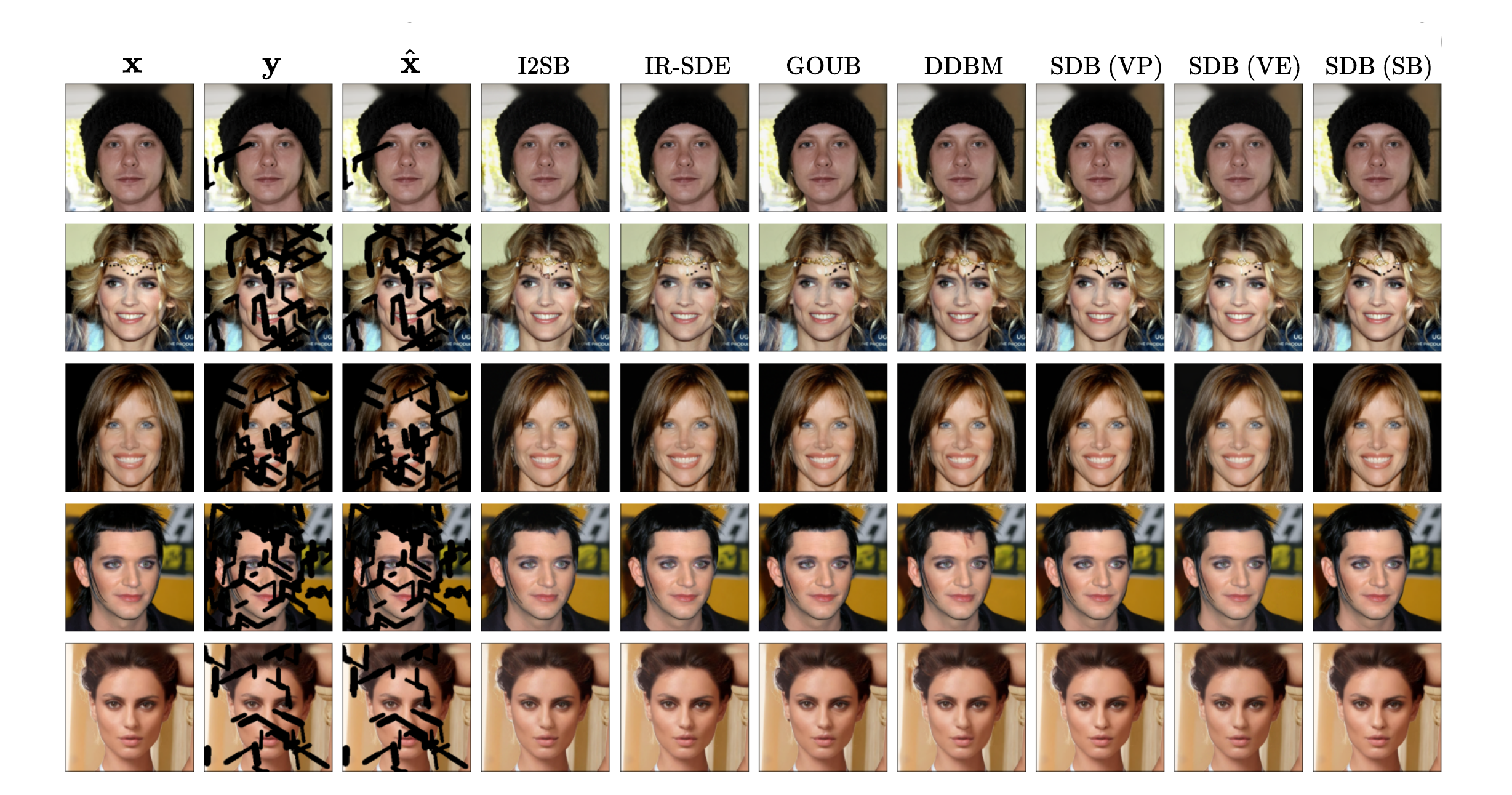}
    \caption{Qualitative comparison of SDB variants with the best-performing baselines (bridge methods). Rows depict the results for inpainting.}
    \label{fig:qual_eval_celebahq}
\end{figure}

\begin{figure}[h]
    \hspace{-2.5em}
    \includegraphics[width=1.12\linewidth]{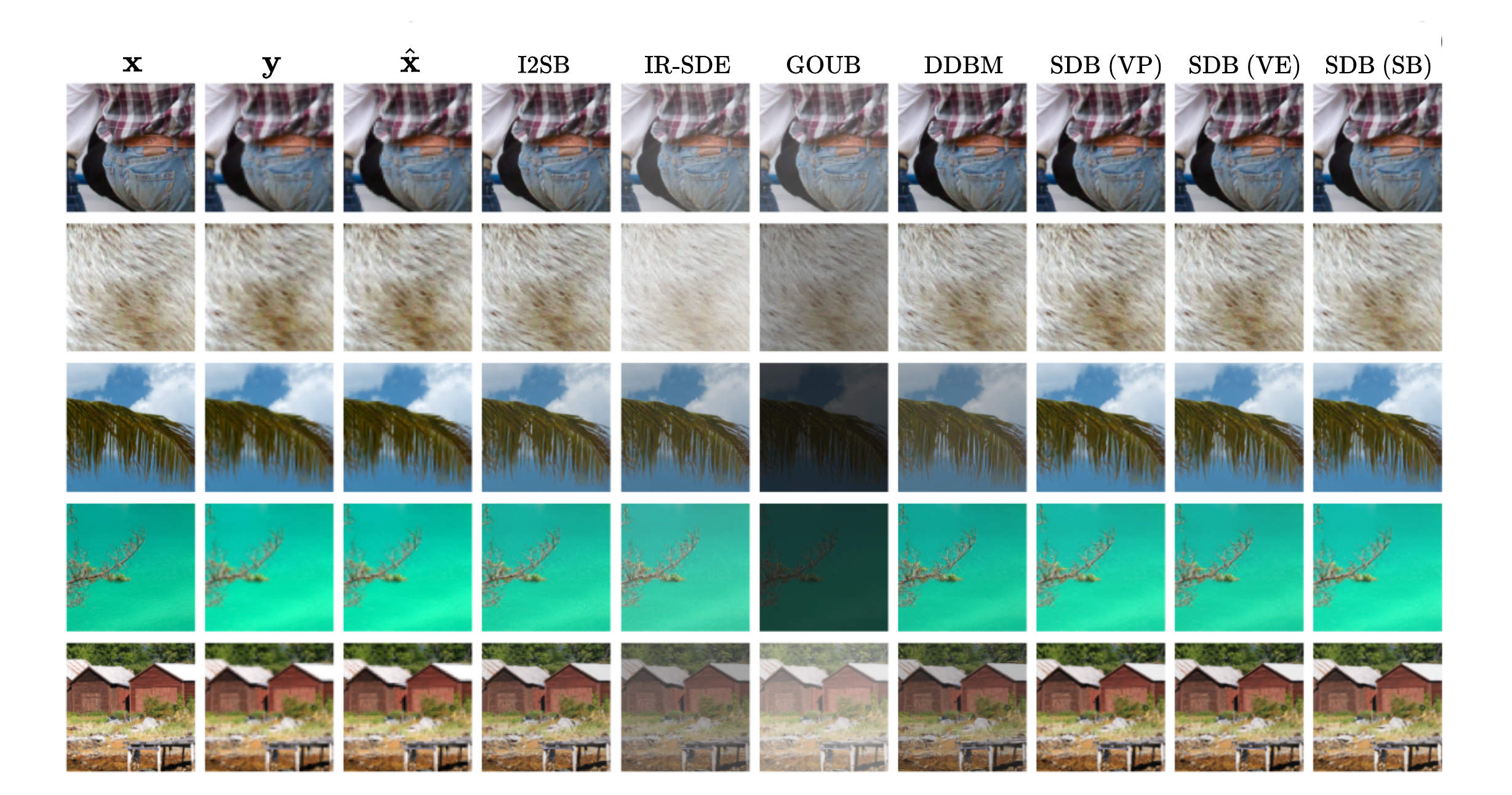}
    \caption{Qualitative comparison of SDB variants with the best-performing baselines (bridge methods). Rows depict the results for superresolution.}
    \label{fig:qual_eval_div2k}
\end{figure}

\begin{figure}[h]
    \hspace{-2.5em}
    \includegraphics[width=1.12\linewidth]{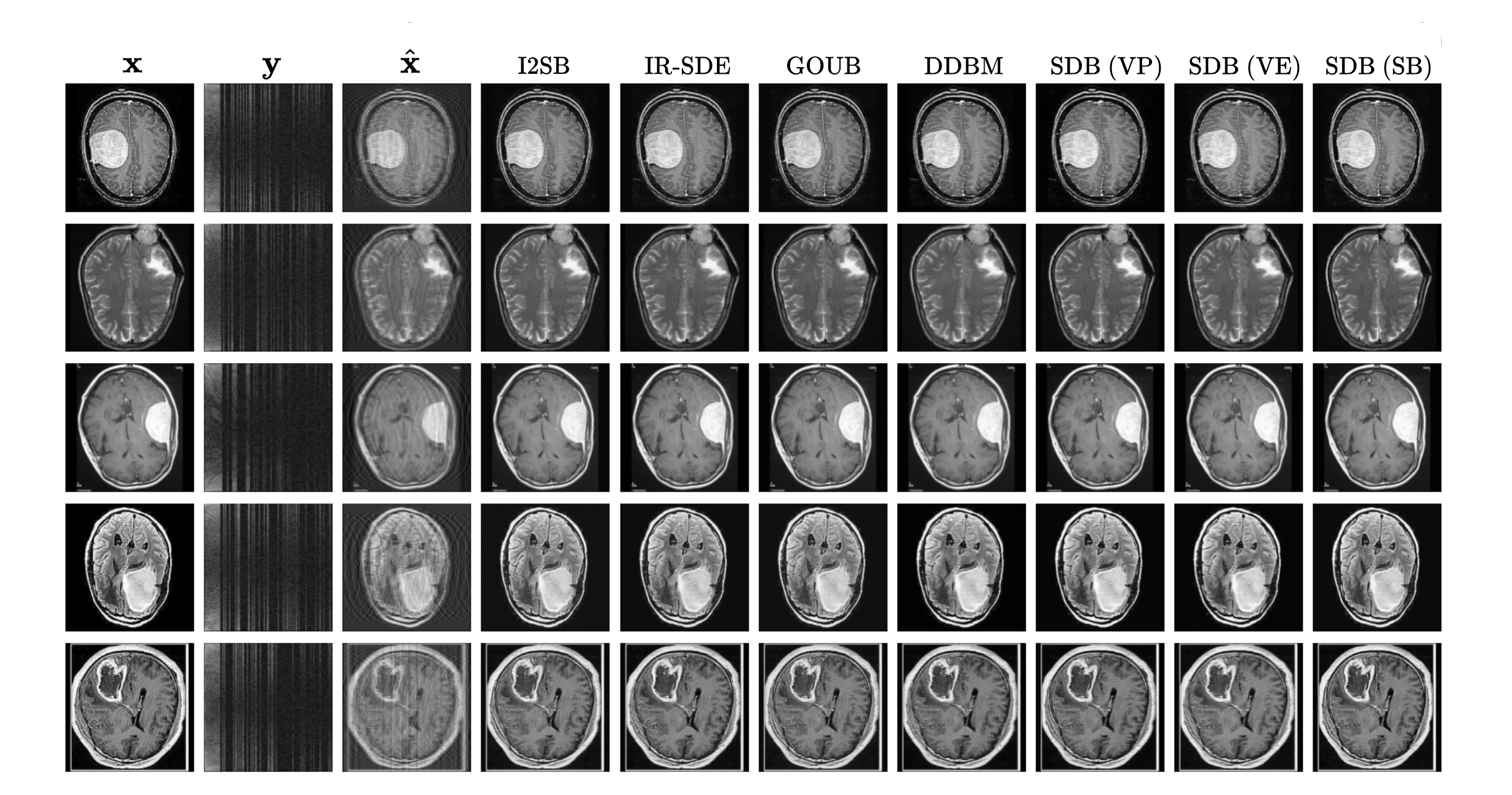}
    \caption{Qualitative comparison of SDB variants with the best-performing baselines (bridge methods). Rows depict the results for MRI reconstruction.}
    \label{fig:qual_eval_brainmri}
\end{figure}

\begin{figure}[h]
    \hspace{-2.5em}
    \includegraphics[width=1.12\linewidth]{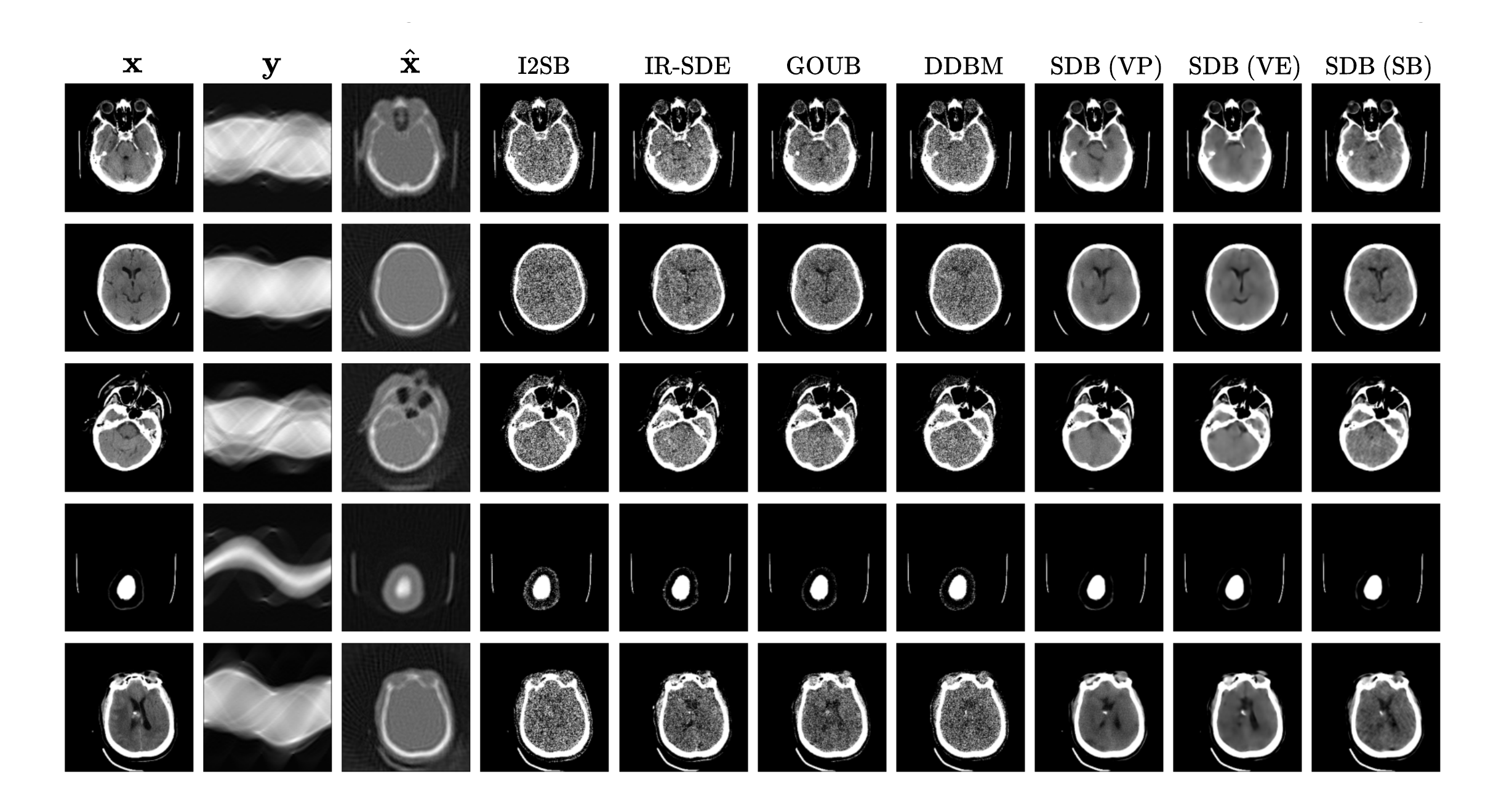}
    \caption{Qualitative comparison of SDB variants with the best-performing baselines (bridge methods). Rows depict the results for CT reconstruction.}
    \label{fig:qual_eval_rsna}
\end{figure}

We provide additional qualitative samples for inpainting (\cref{fig:qual_eval_celebahq}), superresolution (\cref{fig:qual_eval_div2k}), MRI reconstruction (\cref{fig:qual_eval_brainmri}) and CT reconstruction (\cref{fig:qual_eval_rsna}).

\end{document}